\documentclass[sn-mathphys]{sn-jnl}

\jyear{2023}%

\theoremstyle{thmstyleone}%
\theoremstyle{thmstyletwo}%

\theoremstyle{thmstylethree}%

\usepackage{colortbl}

\usepackage{enumitem}

\usepackage{arydshln} 
\usepackage{bm} 
\usepackage{xcolor}

\usepackage[utf8]{inputenc} 
\usepackage[T1]{fontenc}    
\usepackage{hyperref}       
\usepackage{url}            
\usepackage{booktabs}       
\usepackage{amsfonts}       
\usepackage{nicefrac}       
\usepackage{microtype}      
\usepackage{soul}  
\usepackage{amsmath,amsthm,amssymb,amsfonts}

\usepackage{algorithm, algorithmicx, algpseudocode}

\begin{document}


\title{Dynamic Datasets and Market Environments for Financial Reinforcement Learning}


\author[1]{Xiao-Yang Liu}
\author[1]{Ziyi Xia}
\author[1]{Hongyang Yang}
\author[2]{Jiechao Gao}
\author[3]{Daochen Zha}
\author[4]{Ming Zhu}
\author[5]{Christina Dan Wang}
\author[6]{Zhaoran Wang}
\author[7]{Jian Guo}

\affil[1]{Columbia University, USA}
\affil[2]{University of Virginia, USA}
\affil[3]{Rice University, USA}
\affil[4]{SIAT, Chinese Academy of Sciences, China}
\affil[5]{New York University (Shanghai), China}
\affil[6]{Northwestern University, USA}
\affil[7]{IDEA Research, International Digital Economy Academy, China}









\abstract{

The financial market is a particularly challenging playground for deep reinforcement learning due to its unique feature of dynamic datasets. Building high-quality market environments for training financial reinforcement learning (FinRL) agents is difficult due to major factors such as the low signal-to-noise ratio of financial data, survivorship bias of historical data, and model overfitting. In this paper, we present FinRL-Meta, a data-centric and openly accessible library that processes dynamic datasets from real-world markets into gym-style market environments and has been actively maintained by the AI4Finance community. First, following a DataOps paradigm, we provide hundreds of market environments through an automatic data curation pipeline. Second, we provide homegrown examples and reproduce popular research papers as stepping stones for users to design new trading strategies. We also deploy the library on cloud platforms so that users can visualize their own results and assess the relative performance via community-wise competitions. Third, we provide dozens of Jupyter/Python demos organized into a curriculum and a documentation website to serve the rapidly growing community. The open-source codes for the data curation pipeline are available at \url{https://github.com/AI4Finance-Foundation/FinRL-Meta}

}




\maketitle

\section{Introduction}
\label{sec1}

Financial reinforcement learning (FinRL) \cite{liu2021finrl,hambly2021recent} is a promising interdisciplinary field of finance and reinforcement learning, driven by the spirit of ``\textit{trade with an (technology) edge}''\footnote{Find a (technology) edge and position to win.}. In the past decade, deep reinforcement learning (DRL) \cite{sutton2018reinforcement}, as a disruptive technology, has delivered a superhuman performance in Atari games \cite{DQN}, Go \cite{silver2016alphaGo1,silver2017alphaGo2}, StarCraft II \cite{vinyals2019grandmaster}, the recent eye-catching ChatGPT \cite{ouyang2022training}, and GPT-4 \cite{GPT4}. The financial market is a particularly challenging playground for DRL due to the unique feature of \textit{dynamic dataset}, a sharp contrast to the static ImageNet dataset \cite{deng2009imagenet}.

The static ImageNet dataset \cite{deng2009imagenet} initiated the field of deep learning.  Researchers designed and applied many deep neural network models to real-world visual applications. However, these applications deal with ``static'' datasets, such as MNIST, CIFAR-10, Yale Face Database, which are quite different from the financial market, where the data is dynamic in nature. The market trend, development of companies, and economic situation of countries are refreshing over time. 
To handle time-sensitive financial data, models have to learn up-to-date information and adapt to real-time market situations.

Existing works \cite{lussange2021modelling, liu2021finrl, pricope2021deep} have already applied various DRL algorithms in financial applications, including investigating market fragility \cite{raberto2001agent}, designing profitable strategies \cite{liu2018practical, yang2020deep,zhang2020deep}, and assessing portfolio risk \cite{lussange2021modelling, bao2019multiagent}. Exemplar financial trading tasks include
\begin{itemize}
    \item Fundamentals analysis: e.g., value investing, growth investing 
    \item Technical analysis: commodity trading advisor (CTA), momentum, trend following, etc.
    \item Macro strategies: e.g., bonds, gold, crude oil, forex.
    \item Quantitative strategies: statistical arbitrage, merge/event arbitrage, etc.
    \item High-frequency trading: many different sub-strategies inside.
\end{itemize}
Some recent works \cite{lussange2021modelling, amrouni2021abides, market_simulator} have shown that DRL can deliver better trading performance than classical strategies and conventional machine learning methods do on these tasks regarding cumulative return and Sharpe ratio.
However, these works are difficult to reproduce. Several recent efforts have been dedicated to facilitating reproducibility. The FinRL library \cite{liu2020finrl,liu2021finrl} provided an open-source framework for financial reinforcement learning. Unfortunately, it only focused on the reproducibility of backtesting performance by providing several market environments. A conference version of FinRL-Meta \cite{liu2022finrlmeta} provided a financial dataset and benchmark, but it did not provide a dynamic dataset, sentiment data, and market simulator.

However, building near-real market environments for financial reinforcement learning (FinRL) is difficult due to major factors such as low signal-to-noise ratio (SNR) of financial data, partial observation, reward delay, survivorship bias of historical data, and model overfitting. Such a \textit{simulation-to-reality gap} \cite{DulacArnold2020AnEI,dulac2019challenges} degrades the performance of DRL strategies in real markets. A good backtest performance does not necessarily reflect the actual trading performance since the financial dataset could be susceptible to flaws like missing data, noise, and anomalies.
 
Data-centric AI \cite{whang2023data, zha2023data, zha2023data2} is a new trend that appeals to shift the focus from model to data. It focuses on the systematic engineering of data in building AI systems, which can lead to better model behaviors in the real world~\cite{zha2023data}. 
When training DRL agents, if the data is susceptible to flaws, even if our model/strategy can have a good backtest performance, it will be hard to tell whether the model can actually perform well in the real deployment or is just a result of overfitting the dataset. Therefore, building a standard workflow to ensure data quality is imperative to foster reliable market environments and benchmarks and motivate the research and industrialization of FinRL.

In this paper, we present FinRL-Meta, a data-centric and openly accessible library that has been actively maintained by the AI4Finance community, with a particular focus on data quality. The aim of FinRL-Meta is to create an infrastructure to enable real-time paper trading and facilitate the real-world adoption of FinRL technology. This contributes to the broader RL or ML research community since it allows researchers to test DRL agents in real and dynamic environments. 


\begin{figure}[t]
\centering
\includegraphics[scale = 0.06]{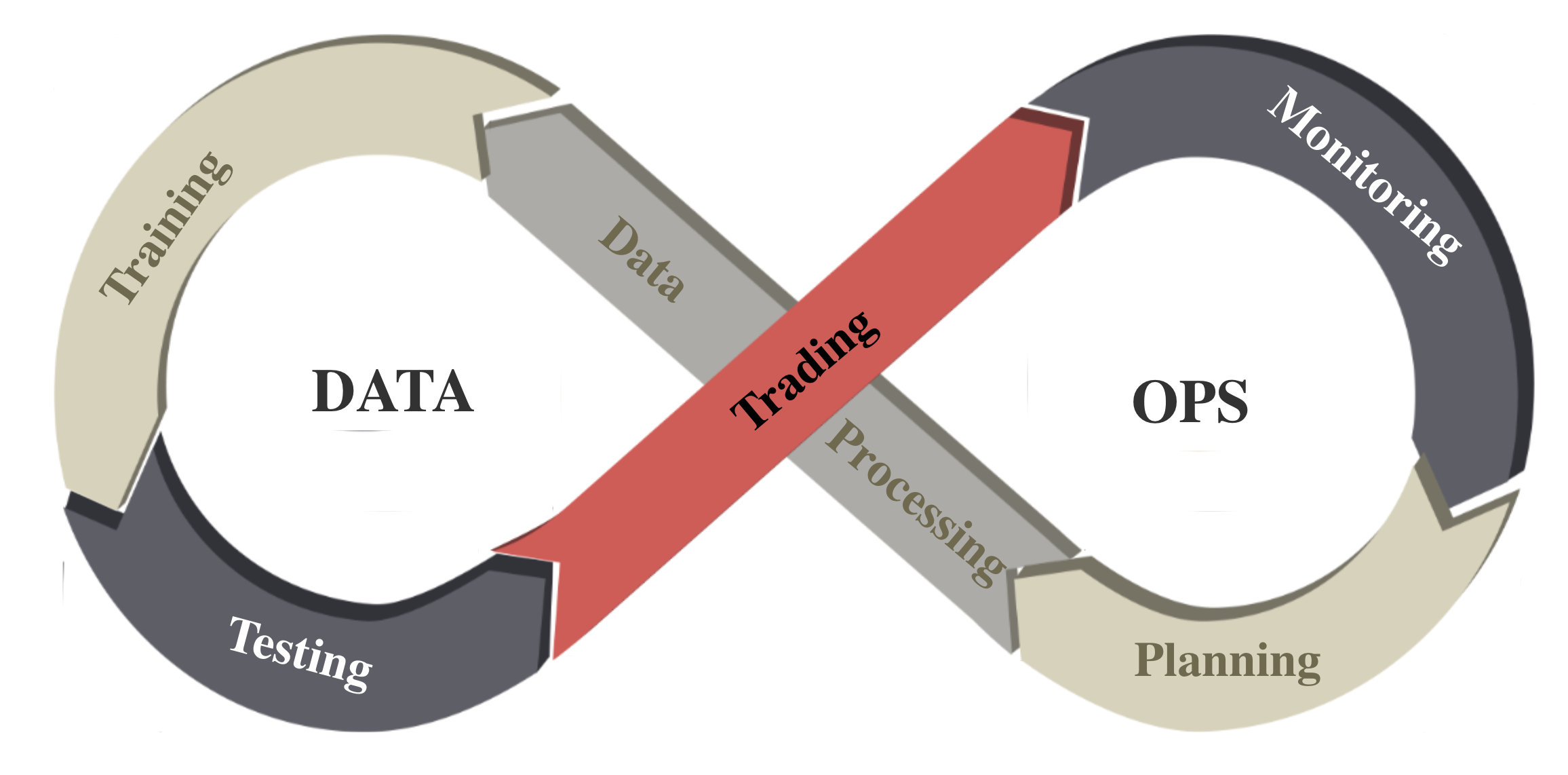}
\hspace{5mm}
\includegraphics[scale = 0.13]{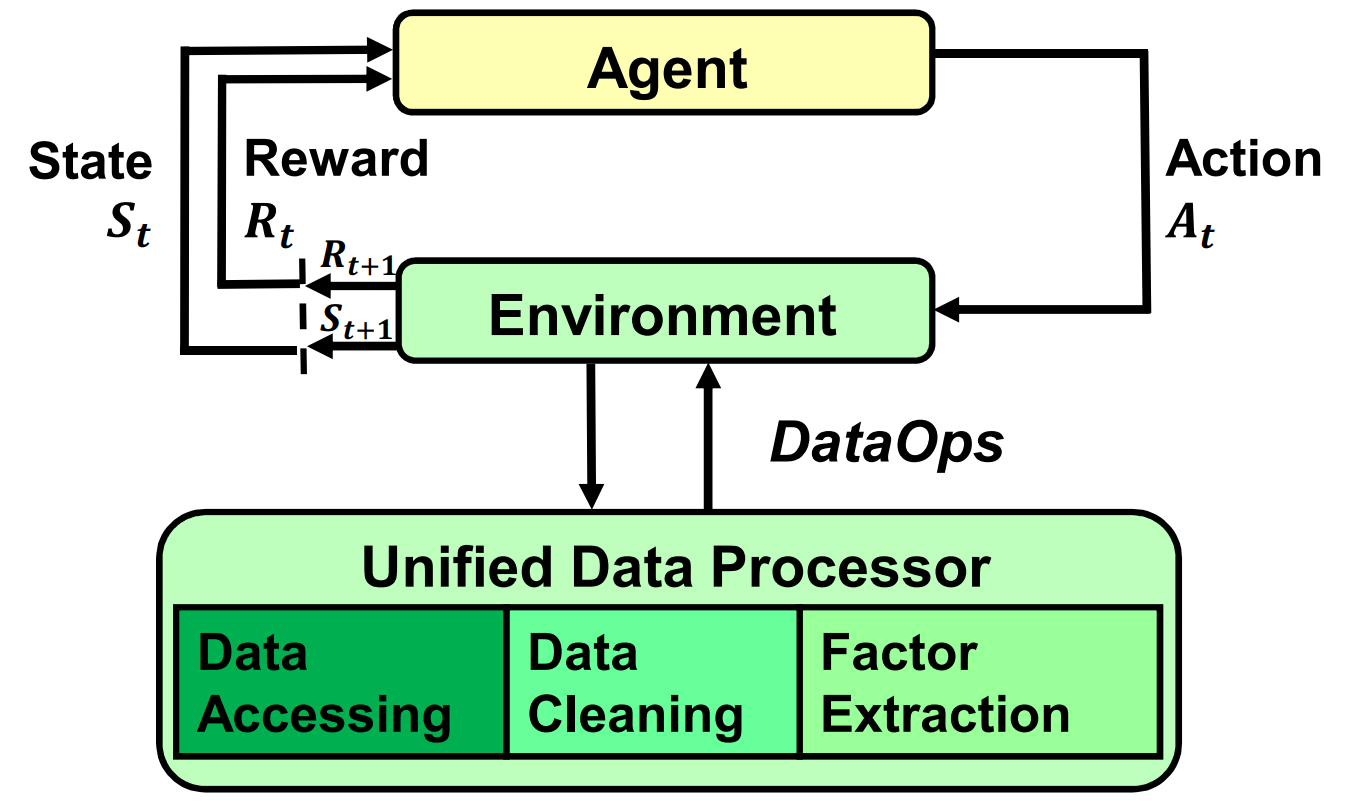}
\caption{DataOps paradigm (left) and FinRL-Meta (right).}
\label{fig:conventional vs neofinrl}
\end{figure}

To handle highly unstructured financial big data, we follow the \textsf{DataOps} paradigm and implement an automatic data curation pipeline in Fig. \ref{fig:conventional vs neofinrl} (left). The \textsf{DataOps} paradigm \cite{atwal2019practical, ereth2018dataops} refers to a set of practices, processes, and technologies that combines automated data engineering and agile development \cite{ereth2018dataops}. It helps reduce the cycle time of data engineering and improve data quality. Following the \textsf{DataOps} paradigm, we design an RLOps pipeline tailored for FinRL to continuously produce DRL benchmarks on dynamic market datasets. The RLOps pipeline consists of the following steps:



\begin{itemize}
    \item The first step is task planning, such as stock trading, portfolio allocation, and cryptocurrency trading.
    \item Then, we do data processing, including data accessing and cleaning, and feature engineering.
    \item Next step is where RL takes part in. In particular, the training-testing-trading process, detailed in Fig. \ref{fig:finrl-meta timeline}.
    \item The final step is performance monitoring.
\end{itemize}



Fig. \ref{fig:conventional vs neofinrl} (right) shows the overview of FinRL-Meta.  
First, following the DataOps paradigm \cite{atwal2019practical, ereth2018dataops}, we provide hundreds of market environments through an automatic data curation pipeline that collects dynamic datasets from real-world markets and processes them into standard gym-style market environments.
Second, we reproduce popular papers as benchmarks, including high-frequency stock trading, cryptocurrency trading and stock portfolio allocation, serving as stepping stones for users to design new strategies. With the help of the data curation pipeline, we hold our benchmarks on cloud platforms so that users can visualize their own results and assess the relative performance via community-wise competitions. 
Third, we provide dozens of Jupyter/Python demos as educational materials, organized in a curriculum for community newcomers with different levels of proficiency and learning goals. At the same time, we maintain a documentation website to serve the rapidly growing community. 

The remainder of this paper is organized as follows. Section 2 describes related works. Section 3 presents an overview of FinRL and the FinRL-Meta framework. Section 4 describes the automatic data curation pipeline of FinRL-Meta. In Section 5, we present several homegrown examples using FinRL-Meta. Finally, we conclude this paper and discuss future works in Section 6.

\section{Related Works}

We review the technology landscape and existing works on financial big data, data-centric AI, DataOps practices, data-driven RL, and FinRL applications.

\subsection{Handling Financial Big Data}

\textbf{Financial big data}: Financial big data refers to the vast amount of data that is available in the financial industry from various sources. By analyzing this data, traders can make informed decisions about investments. Like all big data, financial big data also shares the four key properties, known as the 4V’s\footnote{The Four V’s of Big Data: \url{https://opensistemas.com/en/the-four-vs-of-big-data/}}: volume, variety, velocity, and veracity.
\begin{itemize}
\item 1) Volume. Financial big data has a large scale. Now, the market can support high-frequency trading at the level of microseconds. With high-frequency trading, billions of shares can be traded within a day, generating an enormous amount of data that records these transactions.
\item 2) Velocity. As the market refreshes at the microsecond level, the velocity of data transmission and processing is very important. Companies are continually seeking ways to speed up data transmission and processing to minimize delays and errors. This includes utilizing closer physical proximity to data vendors, improving cable materials, and developing more efficient algorithms.

\item 3) Variety. There are structured and unstructured data in finance. A data vendor may provide structured data for users to access, usually the volume-price data. Besides that, there are a huge amount of alternative data from news, social media, and financial reports that are considered during the process of generating indicators.

\item 4) Veracity. Data quality and availability are extremely important. However, big data's larger scale also brings a latent risk of lack of veracity. Due to its close relationship with money and assets, data quality is particularly sensitive in the finance industry.
\end {itemize}

\textbf{Data-centric AI:} With the 4V's properties, the quality of financial data plays a critical role in enabling strong FinRL models in real deployments. In the past, FinRL research was conducted in a model-centric way, with an emphasis on improving model designs using predetermined datasets. However, solely depending on static datasets does not necessarily result in satisfactory model performance in real-world scenarios, especially when the dataset is flawed~\cite{mazumder2022dataperf}. Furthermore, neglecting the importance of data quality can trigger data cascades~\cite{sambasivan2021everyone}, leading to reduced accuracy in real deployments.

Recently, the attention of researchers and practitioners has gradually shifted toward data-centric AI~\cite{whang2023data, zha2023data, zha2023data2}, with a stronger emphasis on improving data quality by the systematic engineering of data. The benefits of data-centric AI have been validated by both researchers and practitioners~\cite {zha2023data,polyzotis2021can}. In order to drive tangible advancements in FinRL research and deployment, we have made our library data-centric by building upon dynamic datasets and implementing an automated data curation pipeline for quality control.

\textbf{DataOps practices}: From another perspective, a standardized development cycle is necessary for effectively handling highly unstructured financial big data. DataOps \cite{ereth2018dataops, atwal2019practical} applies the ideas of lean development and DevOps to the data science field. DataOps practices have been developed in companies and organizations to improve the quality and efficiency of data analytics \cite{atwal2019practical}. These implementations consolidate various data sources, and unify and automate the pipeline of data analytics, including data accessing, cleaning, analysis, and visualization.

The DataOps paradigm~\cite{ereth2018dataops}, or more accurately the methodology, is a way of organizing people, processes and technology to deliver reliable and high-quality data efficiently to all its users. The practice of DataOps focuses on enabling collaboration across the organization to drive agility, speed of delivery and new data initiatives. By leveraging the power of automation, DataOps aims to address the challenges associated with inefficiencies in access, preparation, integration and availability of data. 

Many researchers studied FinRL applications \cite{liu2018practical, yang2020deep, zhang2020deep, ardon2021towards, amrouni2021abides, coletta2021towards} by building their own market environments. Despite the above-mentioned open-source libraries that provide some valuable settings, there are no established benchmarks yet. On the other hand, the data accessing, cleaning and feature/factor extraction processes are usually limited to data sources like Yahoo Finance and Wharton Research Data Services (WRDS).

However, the DataOps methodology has not been applied to FinRL research. Most researchers access data, clean data, and extract technical indicators (features) in a case-by-case manner, which involves heavy manual work and may not guarantee high data quality. 

\subsection{Data-Driven Reinforcement Learning}

\textbf{Data-driven RL}: If a policy is learned from a collected dataset, it is promising that we can get the data-driven strategy\footnote{Note that ``data-driven'' and ``data-centric'' are two distinct concepts. The former refers to utilizing data to guide policy training, whereas the latter means placing data quality in the central role in FinRL development. The endeavors of ``data-driven'' and ``data-centric'' approaches complement each other in their efforts to enhance overall policy performance.}. Based on the datasets, RL will be a powerful method to perform data-driven strategies without any interaction with the environment or human intervention. RL will greatly reduce human labor and therefore improves automation. 


\textbf{Offline RL \cite{levine2020offline}}: Offline RL is a typical data-driven formulation of reinforcement learning problems. In offline RL, agents learn behaviors from a fixed dataset, without the process of exploration in the environment. It has great potential in tasks where collecting real-time data is inconvenient, either too expensive or risky(e.g., robotic tasks, autonomous driving), or the amount of data is limited (e.g., stock market, clinical surgery).

\textbf{Curriculum learning}: Curriculum learning is a technique that trains the model using multiple stages from simple to complex. With fine-tuning for specific tasks, curriculum learning could have faster convergence and find better minima. \cite{jpmorgan2023asset} proposed to use two stages of training for the portfolio management task, first using a neural network to fit the mean-variance optimization, then fine-tuning the model with online reinforcement learning. Well-known products like AlphaGo and ChatGPT both use similar techniques.

\textbf{RL from human feedback (RLHF)}: The idea of RLHF was first introduced by OpenAI and DeepMind in 2017 \cite{christiano2017deep}. RLHF allows human feedback as part of the agent's reward function, making it a better alignment of model performance and human expectations. One of the main reasons that ChatGPT outperforms other large language models is its appropriate usage of RLHF.

Next, we review popular projects:
\begin{itemize}[leftmargin=*]
    \item \textbf{OpenAI gym} \cite{brockman2016openai} Environments are crucial for training DRL agents \cite{sutton2018reinforcement}. OpenAI gym provides standardized environments for a collection of benchmark problems that expose a common interface, which is widely supported by many libraries \cite{stable-baselines, liang2018rllib, elegantrl}. Three trading environments, \textsf{TradingEnv, ForexEnv, and StocksEnv}, are included to support stock and FOREX markets. However, it has not been updated for years.
    \item \textbf{Game of Go.} AlphaGo \cite{silver2016alphaGo1} and AlphaGo Zero \cite{silver2017alphaGo2} are programs for games of Go. AlphaGo combines Monte Carlo simulation with value and policy networks, and becomes the first computer program that defeats world champions in Go game. AlphaGo Zero is the updated version, it learns by reinforcement learning by playing against itself, without extra human data or knowledge. These programs also provide suggestions for financial reinforcement learning, e.g., how to train the policy network by supervised learning and self-play. 
    \item \textbf{D4RL} \cite{fu2020d4rl} introduces the idea of \textit{datasets for deep data-driven reinforcement learning} (D4RL). It provides benchmarks in offline RL. However, D4RL does not provide financial environments.
    \item \textbf{FinRL} \cite{liu2020finrl,liu2021finrl} is an open-source library that builds a full pipeline for financial reinforcement learning. It contains three market environments, i.e., stock trading, portfolio allocation, and crypto trading, and two data sources, i.e., Yahoo Finance and WRDS. However, those market environments of FinRL cannot meet the community's growing demands.
    \item \textbf{NeoRL} \cite{qin2021neorl} collected offline RL environments for four areas, CityLearn \cite{vazquez2019cityLearn}, FinRL \cite{liu2020finrl,liu2021finrl}, Industrial Benchmark \cite{hein2017benchmark}, and MuJoCo \cite{todorov2012mujoco}, where each area contains several gym-style environments. Regarding financial aspects, it directly imports market environments from FinRL. 
    \item \textbf{ChatGPT} \cite{ouyang2022training} and GPT-4 \cite{GPT4} are large language models for dialogues, which can be used as the data source and feature/factor calculation for data-driven financial reinforcement learning. However, it is not as professional as we think; therefore, how to train as an intelligent advisor is an important problem. We can follow several steps using dialogue, including guides, listing examples, admitting mistakes, etc. 
\end{itemize}

As a new rising technology, RL is a promising tool for complicated financial tasks. Recently, many Wall Street companies have shown great interest in this rising technology. However, RL's instability makes it hard to put into practice in the industry.
AlphaGo and ChatGPT's success in data-feed multi-stage learning could be a great approach to adopting RL into their workflow. Hedge funds first feed their private data into supervised learning (imitation learning) models for training, and then use reinforcement learning to fine-tune the model based on interaction with the market environments to achieve super-human performance.

\subsection{Financial Reinforcement Learning (FinRL)} \label{sec_RLOps}

\begin{figure}[t]
\centering
\includegraphics[width=4.5in]{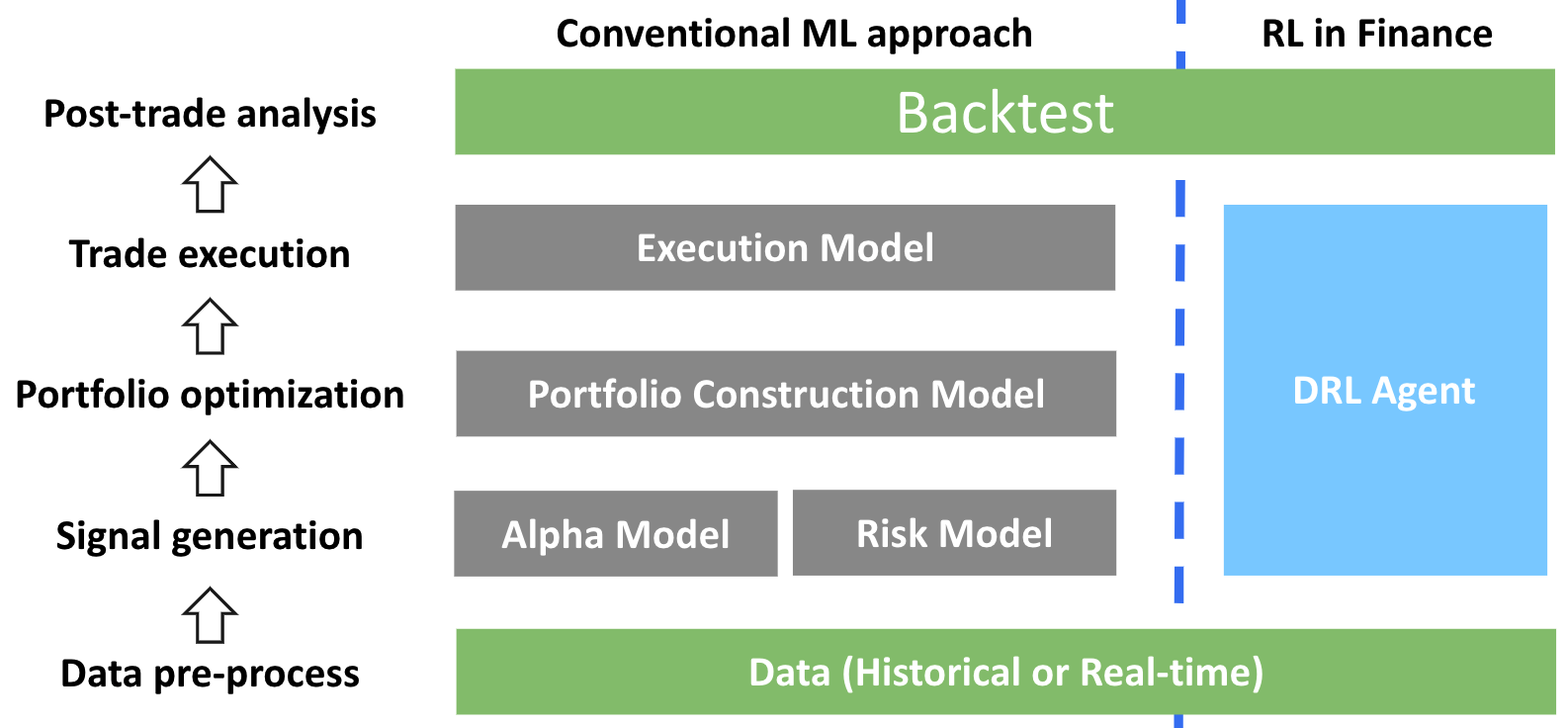}
\caption{Comparison between conventional machine learning approach and RLOps in finance for an algorithmic trading process. 
}
\label{fig_structure}
\end{figure}

\textbf{RLOps paradigm in finance}. Algorithmic trading \cite{Treleaven2013AlgorithmicTR, Nuti2011AlgorithmicT} has been widely adopted in financial investments. The lifecycle of a conventional machine learning strategy may include five general stages, as shown in Fig. \ref{fig_structure} (left), namely data pre-processing, modeling and trading signal generation, portfolio optimization, trade execution, and post-trade analysis. Recently, deep reinforcement learning (DRL) \cite{silver2016alphaGo1, silver2017alphaGo2, sutton2018reinforcement} has been recognized as a powerful approach for quantitative finance, since it has the potential to overcome some important limitations of supervised learning, such as the difficulty in label specification and the gap between modeling, positioning, and order execution.

We would like to extend the principle of \textit{MLOps} \cite{alla2021mlops}\footnote{MLOps is an ML engineering culture and practice that aims at unifying ML system development (Dev) and ML system operation (Ops).} to the \textit{RLOps in finance} paradigm that implements and automates the continuous training (CT), continuous integration (CI), and continuous delivery (CD) for trading strategies. Such a paradigm will have vast profit potential from a broadened horizon and fast speed, which is critical for wider DRL adoption in real-world financial tasks. The \textit{RLOps in finance} paradigm, as shown in Fig. \ref{fig_structure} (right), integrates middle stages (i.e., modeling and trading signal generation, portfolio optimization, and trade execution) into a DRL agent. Such a paradigm aims to help quantitative traders develop an end-to-end trading strategy with a high degree of automation, which removes the latency between stages and results in a compact software stack. The major benefit is that it can explore the vast potential profits behind the large-scale financial data, exceeding the capacity of human traders; thus, the trading horizon is lifted into a potentially new dimension. Also, it allows traders to continuously update trading strategies, which equips traders with an edge in a highly volatile market. However, the large-scale financial data and fast iteration of trading strategies bring imperative challenges in terms of computing power.

\textbf{FinRL applications}: Along with the RLOps paradigm, different FinRL applications can be constructed and deployed. Electronic trading is popular in many countries and is used in stock exchanges, electronic order books, over-the-counter markets, foreign exchange, etc. Electronic trading enhances liquidity since traders can easily buy or sell assets. We describe several FinRL applications inspired by a complete survey \cite{hambly2021recent}, including optimal execution, portfolio optimization, option pricing and hedging, market making, smart order routing, and robo-advising.
\begin{itemize}[leftmargin=*]
\item 1) Optimal execution is the problem of maximizing the return from buying or selling a given amount of an asset within a given time period. A classical framework is the Almgren–Chriss model, which relies heavily on the assumptions of the dynamics and the permanent and temporary price impact. There are several popular criteria to evaluate the performance of execution strategies, e.g., the profit and loss, implementation shortfall, and the Sharp ratio. 
\item 2) Portfolio optimization aims to maximize some objective function by selecting and trading the best portfolio of assets. One typical model is mean-variance portfolio optimization, which aims to maximize the return for a given risk measured by variance. 
\item 3) Option pricing and hedging are important in finance. The Black-Scholes model is a typical mathematical model which aims to get the price of a European option given several constraints: stock price, expiration time, and the payoff at expiry. 
\item 4) Market making aims to earn the bid-ask spread by providing liquidity to the market by placing buy/sell limit orders in the limit order books. 
\item 5) Robo-advising or automated investment managing provides online financial advice with minimal human intervention. ChatGPT \cite{ouyang2022training} and GPT-4 \cite{GPT4} may be promising advisors if we guide them step by step. 
\end{itemize}


\section{FinRL Tasks and FinRL-Meta Framework}

\definecolor{Gray}{RGB}{217,234,211}
\begin{table}
\caption{List of state space, action space, and reward function.}
\small
\renewcommand{\arraystretch}{1.2}
\centering
\begin{tabular}{|l|l|m{200pt}<{\centering}|}
    \hline   
    \textbf{Key components} & \textbf{Attributes} 
    \\ \hline
    \multirow{4}{2cm}{State} & Balance ${b}_{t}\in \mathbb{R}_+$;~Shares $\bm{h}_{t}\in \mathbb{Z}_+^{n}$ \\
    \cdashline{2-3}[0.8pt/2pt]
    & Opening/high/low/close price $\bm{o}_{t}, \bm{h}_{t}, \bm{l}_{t},\bm{p}_{t} \in \mathbb{R}_+^{n}$ \\ 
    \cdashline{2-3}[0.8pt/2pt]
    & Trading volume $\bm{v}_{t}\in \mathbb{R}_+^{n}$ \\
    \cdashline{2-3}[0.8pt/2pt]
    & Fundamental indicators; Technical indicators \\
    \cdashline{2-3}[0.8pt/2pt]
    & Social data; Sentiment data \\
    \cdashline{2-3}[0.8pt/2pt] 
    & Alpha and beta signals; Smart beta indexes, etc. \\ 
    \hline
     \multirow{2}{2cm}{Action} & Buy/Sell/Hold   \\
    \cdashline{2-3}[0.8pt/2pt]
    & Short/Long \\
    \cdashline{2-3}[0.8pt/2pt]
    & Portfolio weights 
    \\ \hline
     \multirow{3}{2cm}{Reward} &  Change of portfolio value \\  
    \cdashline{2-3}[0.8pt/2pt]
    & Portfolio log-return \\
    \cdashline{2-3}[0.8pt/2pt]
    & Sharpe ratio 
    \\ \hline
     \multirow{3}{2cm}{Environments} &  Dow-$30$, S\&P-$500$, NASDAQ-$100$ \\  
    \cdashline{2-3}[0.8pt/2pt]
    &  Cryptocurrencies \\
    \cdashline{2-3}[0.8pt/2pt]
    & Foreign currency and exchange \\ 
    \cdashline{2-3}[0.8pt/2pt]
    &  Futures;~Options;~ETFs;~Forex \\ 
    \cdashline{2-3}[0.8pt/2pt]
    & CN securities;~US securities \\
    \cdashline{2-3}[0.8pt/2pt]
    & Paper trading; Live Trading
    \\ \hline
\end{tabular}\vspace{0.050in}
\label{event:eventTypes}
\end{table}

We introduce the Markov Decision Process (MDP) as a mathematical model of FinRL tasks, summarize FinRL challenges, and then provide an overview of the proposed FinRL-Meta framework.

\subsection{Modeling Financial Reinforcement Learning (FinRL)} \label{subsec:modeling_finrl}

FinRL tasks in general take the form of sequential decision-making problems, which can be mathematically formulated as a Markov Decision Process (MDP) with five-tuple $(S, A, R, \mathbb{P}, \gamma)$ as follows
\begin{itemize}
    \item State space $S$ consists of all possible states;
    \item Action space $A$ consists of all available actions;
    \item Reward function $R(s,a,s'): S\times A\times S\rightarrow\mathbb{R}$ assigns a real-valued reward to a transition $(s, a, s')$;
    \item Transition probability $\mathbb{P}(s'\vert a,s):S\times A\times S\rightarrow [0,1]$ models the dynamics of a system (a.k.a., environment);
    \item Factor $\gamma \in (0, 1]$ discounts a future reward back to its present value.
\end{itemize}

MDP is a well-formed mathematical model for sequential decision-making tasks in that a decision-maker can partially or fully influence the outcome. State $S$ and action $A$ define the input and output for decision-making tasks, and reward function $R$ allows a model to learn by goal-seeking optimization. Transition probability $\mathbb{P}$ allows the target problem to be stochastic, which is usually closer to reality. The discount factor $\gamma$ makes the model consider the future reward when making decisions.

We summarize the state spaces, action spaces, and reward functions of FinRL applications in Table \ref{event:eventTypes}. States usually demonstrate the condition of market and the assets, such as the balance, shares of stocks, OHLCV data, technical indicators, sentiment data, etc. Actions are the operations allowed in the market, including buy/sell/hold certain shares of the stock, short or long, change of portfolio weights on stocks, etc. The reward functions indicate what kind of objective we want the agent to achieve, for example, a change of portfolio value (a larger positive change leads to a larger positive reward, vice versa) for maximizing excess return, Sharpe ratio for balancing return and risk, etc.

\textbf{Objective function}: Many financial tasks can be written in the form of optimization problems. When using RL, the objective function is to find a policy $\pi(s,a)$ that maximizes the discounted cumulative return $r=\sum_{t=0}^{T}\gamma^t R(s_t,a,s_{t+1 })$.

\textbf{Example I}. For the stock trading task \cite{liu2018practical} on $30$ constituent stocks of the Dow Jones Industrial Average (DJIA) index, we specify the ``state-action-reward'' as follows:
\begin{itemize}
    \item State $\bm{s_t}=[b_t,\bm{p_t},\bm{f_t},\bm{h_t}] \in \mathbb{R}^{30(I+2)+1}$, where scalar $b_t\in \mathbb{R}_+$ is the remaining balance in the account, $\bm{p_t}\in \mathbb{R}_+^{30}$ is the prices of $30$ stocks, $\bm{f_t}\in \mathbb{R}^{30\cdot I}$ is a feature vector and each stock has $I$ technical indicators, and $\bm{h_t}\in \mathbb{R}_+^{30}$ denotes the share holdings, where $\mathbb{R}_+$ is the set of non-negative real numbers.
    \item Action $\bm{a_t} \in \mathbb{R}^{30}$ denotes the trading operations on the 30 stocks, i.e., $\bm{h_{t+1}}=\bm{h_t} + \bm{a_t}$. When an entry $\bm{a}_t^i > 0, i=1, ..., 30$, it means a buy-in of $\bm{a}_t^i$ shares on the $i$-th stock, negative action $\bm{a}_t^i < 0$ for selling, and zero action $\bm{a}_t^i = 0 $  keeps $\bm{h}_t^i$ unchanged.
    \item Reward function $R(\bm{s_t}, \bm{a_t}, \bm{s_{t+1}})\in \mathbb{R}$: Reward is an incentive signal to encourage the trading agent to execute action $\bm{a_t}$ at state $\bm{s_t}$. In the stock trading task \cite{liu2018practical}, the reward function is set to be the change of total asset values, i.e., $R(\bm{s_t}, \bm{a_t}, \bm{s_{t+1}})=v_{t+1}-v_t$, where $v_t$ and $v_{t+1}$ are the total asset values at state $\bm{s_t}$ and $\bm{s_{t+1}}$, respectively, i.e., $v_t=\bm{p_t}^{\intercal}\bm{h_t}+b_t\in \mathbb{R}$.
\end{itemize}

\textbf{Example II}. Another task is portfolio optimization, also on the 30 constituent stocks of the Dow Johnes Industrial Average (DJIA) index. The "state-action-reward" are specified as follows:
\begin{itemize}
    \item State $\bm{s_t}=[v_t, \bm{p_t},\bm{f_t},\bm{w_t}] \in \mathbb{R}^{30(I+2)+1}$, where $\bm{p_t}\in \mathbb{R}_+^{30}$ is the current prices of $30$ stocks, $\bm{f_t}\in \mathbb{R}^{30\cdot I}$ is a feature vector and each stock has $I$ technical indicators, $\bm{w_t}\in\mathbb{R}^{30}$ is the portfolio weight allocated on last time, and $v_t = v_{t-1}\bm{w_{t-1}}^{\intercal}\bm{p_t}/\bm{p_{t-1}}$ is the total asset value at time $t$.
    \item Action $\bm{a_t} \in \mathbb{R}^{30}$ denotes the new weights assigned to each of the 30 stocks. Note there are always $\sum_{i=1}^{30} \bm{a}_t^i=1$. Each $\bm{a}_t^i>0$ means allocating $a_t^i$ amount of total asset on the $i$-th stock.
    After $\bm{a_t}$ is applied, $\bm{w_{t+1}}$ in the state will record the new $\bm{a_t}$, and $v_{t+1}$ will be recalculated.
    \item Reward function $R(\bm{s_t}, \bm{a_t}, \bm{s_{t+1}})\in \mathbb{R}$: Similar to the stock trading task, the reward of portfolio optimization task is also set to be the change of total asset values.
\end{itemize}

\subsection{FinRL Challenges}

Training and testing environments based on historical data may not simulate real markets accurately due to the \textit{simulation-to-reality gap} \cite{DulacArnold2020AnEI,dulac2019challenges}, and thus a trained agent cannot be directly deployed in real-world markets. We summarize the main  FinRL challenges as follows:
\begin{itemize} [leftmargin=*]
\item \textbf{Low signal-to-noise ratio (SNR)}: Data from different sources may contain large noise \cite{wilkman2020feasibility} such as random noise, outliers, etc. It is challenging to identify alpha signals or build smart beta indices using noisy datasets.
\item \textbf{Survivorship bias of historical market data}: Survivorship bias is caused by a tendency to focus on existing stocks and funds without consideration of those that are delisted \cite{brown1992survivorship}. It could lead to an overestimation of stocks and funds, which will mislead the agent.
\item \textbf{Model overfitting}: Existing research mainly report backtesting results. It is highly possible that authors are tempted to tune hyper-parameters and retrain the agent multiple times \footnote{There is information leakage.} to obtain better backtesting results, resulting in model overfitting \cite{gort2022deep,de2018advances}. This might lead to big trouble during real-time trading.
\item \textbf{Delay}: Financial markets have delays due to data transmission, reward feedback, and actuation. The true reward may be based on the users' interaction with the markets, which may take several days. As the delay increases, the performance of deep reinforcement learning decreases.
\item \textbf{Partial observation}: The full observability assumption of financial markets can be extended to partial observation (the underlying states cannot be directly observed), i.e., partially observable Markov Decision Process (POMDP). A POMDP model utilizes a Hidden Markov Model (HMM) \cite{mamon2007hidden} to model a time series that is caused by a sequence of unobservable states. 
Considering the noisy financial data, it is natural to assume that a trading agent cannot directly observe market states. Studies suggested that the POMDP model can be solved by using recurrent neural networks, e.g., an off-policy Recurrent Deterministic Policy Gradient (RDPG) algorithm \cite{liu2020adaptive}, and a  long short-term memory (LSTM) network that encodes partial observations into a state of a reinforcement learning algorithm \cite{rundo2019deep}.
\item \textbf{Multi-objective reward function}: When we optimize one metric, some other metrics may need to be constrained or improved. Therefore, a trade-off between these metrics may be required. In addition, the reward formulation involves the weights of metrics, i.e., the weights should be fine-tuned manually. 
\item \textbf{Low interpretability}: In deep reinforcement learning, neural networks are used to fit the Q-value functions and policies. However, neural networks are black-box; therefore, deep reinforcement learning is of low interpretability.

\end{itemize}

\subsection{Overview of FinRL-Meta Framework}

\begin{figure*}
\centering
\includegraphics[scale =0.15]{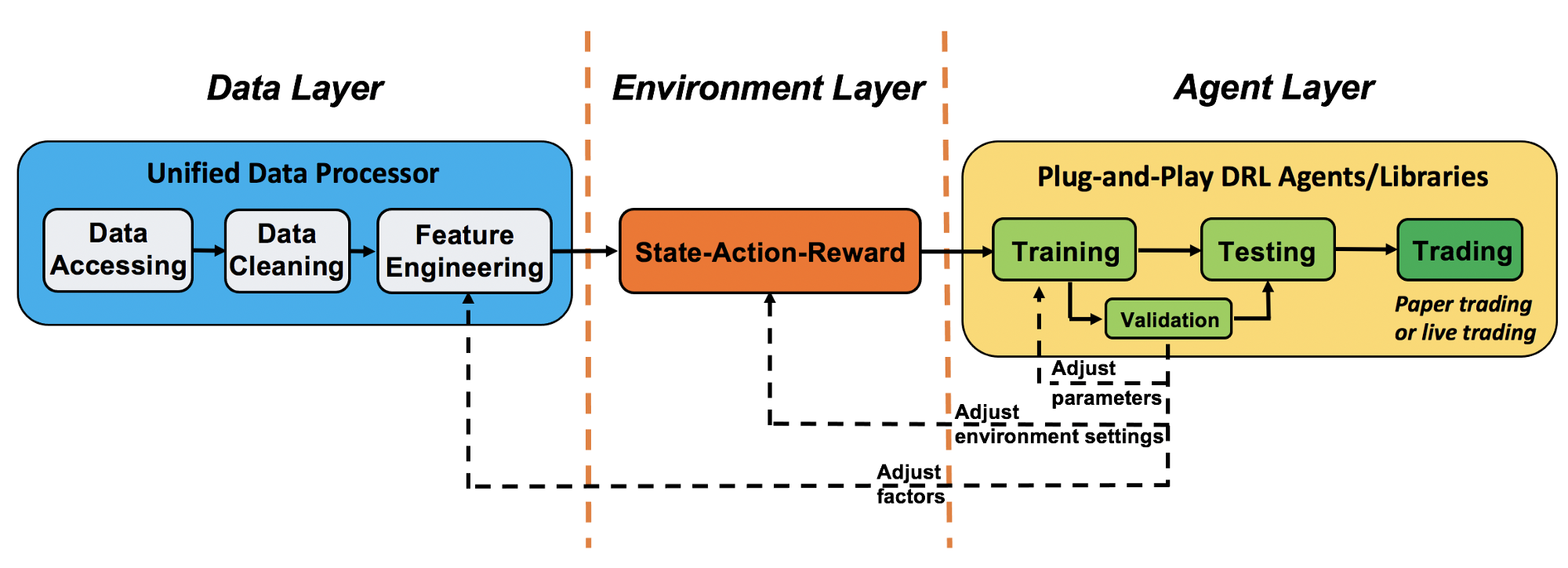}\vspace{-0.1in}
\caption{Overview of FinRL-Meta framework, an automatic data curation pipeline.}
\label{fig:finrl-meta overview}
\vspace{-1mm}
\end{figure*}

Following the DataOps paradigm in Fig. \ref{fig:conventional vs neofinrl} and the standard process of data-centric AI~\cite{zha2023data}, FinRL-Meta builds a universe of market environments for data-driven financial reinforcement learning. FinRL-Meta follows the \textit{de facto} standard of OpenAI Gym \cite{brockman2016openai} and the \textit{lean principle} of software development. We have an automatic pipeline using the dynamic dataset with the following steps: 1). task planning, 2). data processing, 3). training-testing-trading pipeline, 4). performance monitoring.

\subsubsection{Layer Structure and Extensibility}
We adopt a layered structure that consists of three layers, data layer, environment layer, and agent layer, as shown in Fig. \ref{fig:finrl-meta overview}. Layers interact through end-to-end interfaces, achieving high extensibility. For updates and substitutes inside a layer, this structure minimizes the impact on the whole system.  Moreover, the layer structure allows easy extension of user-defined functions and fast updating of algorithms with high performance.

\textbf{Data Layer}: we follow the DataOps paradigm for data curation to reduce the cycle time of data engineering and improve data quality. First, in the data processing step, we provide APIs to collect time series price data and sentiment data from different platforms with a unified interface. Second, the data cleaning step will clean up the raw data, which is usually unstructured and contain different kinds of errors. Third, we will add technical indicators to the data to provide more information on the market in the feature engineering step.

\textbf{Environment Layer}: The well-processed data from the data layer will be made into a gym-style market environment in the environment layer. According to the chosen task, state-action-reward will be set. FinRL-Meta also provides the option of multiprocessing training via vector environment to accelerate the training process.

\textbf{Agent Layer}: we allow a user to plug in a DRL agent and play with a market environment from the environment layer. Currently, three libraries are supported, including Stable-baseline 3, RLlib, and ElegantRL. We call this plug-and-play mode, which will be introduced in detail later.

\begin{figure*}
\centering
\includegraphics[scale = 0.6]{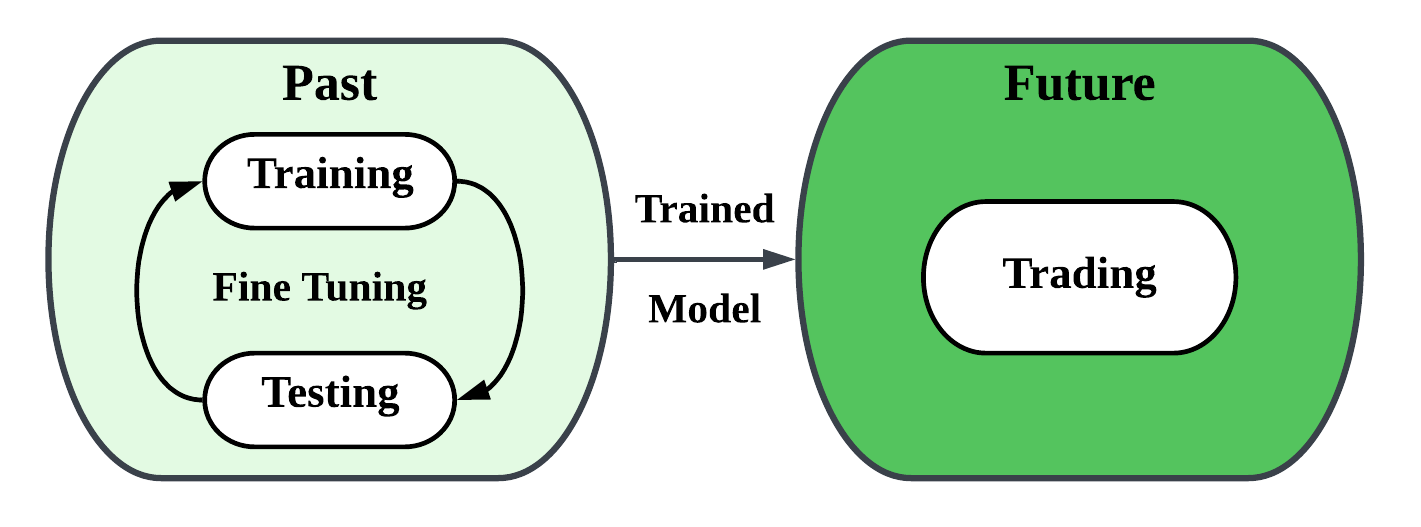}
\caption{Overview of the training-testing-trading pipeline in FinRL-Meta.}

\label{fig:finrl-meta timeline}
\vspace{-1mm}
\end{figure*}

\subsubsection{Dynamic Datasets and Training-Testing-Trading Pipeline}

For dynamic financial data, it is crucial to keep the model learning the latest information from the market. However, it will be very time-consuming to process data and train the model frequently. Thus, FinRL-Meta brings forward the concept of the dynamic dataset.

The dynamic dataset is a standardized workflow of downloading and processing data following the need for a ``training-testing-trading'' pipeline periodically. This makes the time of training controllable.

As shown in Fig. \ref{fig:finrl-meta timeline}, we deploy a training-testing-trading pipeline. The DRL approach follows a standard end-to-end pipeline. The DRL agent is first trained in a training environment and then fined-tuned (adjusting hyperparameters) in a validation environment. Then the validated agent is tested on historical datasets (backtesting). Finally, the tested agent will be deployed in paper trading or live trading markets. Our construction of a dynamic dataset fits the training-testing-trading pipeline well, providing a potential standard workflow of financial reinforcement learning. 

\subsubsection{Plug-and-play Mode for DRL Algorithms}

A DRL agent can be directly plugged in the above training-testing-trading pipeline. The following DRL libraries are supported:
\begin{itemize} [leftmargin=*]
    \item \textbf{ElegantRL \cite{elegantrl}}: Lightweight, efficient and stable algorithms using PyTorch.
    \item \textbf{Stable-Baselines3 \cite{stable-baselines}}: Improved DRL algorithms based on OpenAI Baselines.
    \item \textbf{RLlib \cite{liang2018rllib}:} An open-source DRL library that offers high scalability and unified APIs. 
\end{itemize}

Take the stock trading task described in Section \ref{subsec:modeling_finrl} as an example. We first download and preprocess the historical data of the 30 constituent stocks of the DJIA index. Then, we construct an environment with the state-action-reward specified in Section \ref{sec:datasets}. Given this environment, we choose an algorithm from one of the above three DRL libraries, plug it in with default parameters, and train a trading agent. Users could compare the performances of different algorithms from the same library or different libraries.

\section{Automatic Data Curation Pipeline for Market Environments} \label{sec:datasets}

\begin{figure}
\centering
\includegraphics[scale=0.4]{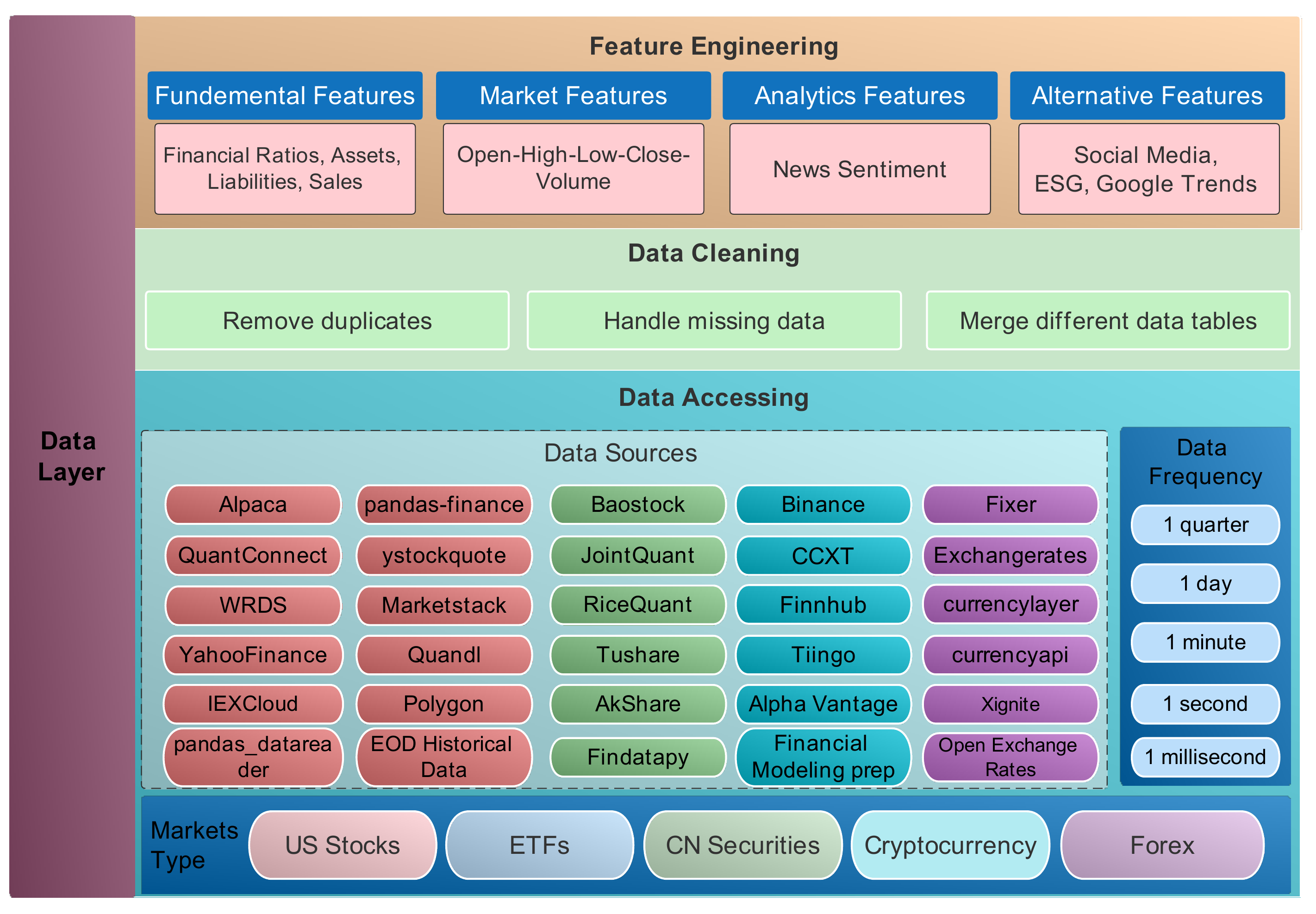}
\caption{Data layer of FinRL-Meta.}
\label{fig:datalayer}
\end{figure}

Financial big data is usually unstructured, which makes data curation necessary. Following the principles of data-centric AI \cite{whang2023data, zha2023data, zha2023data2}, we construct an automatic data curation pipeline to process and engineer data. Specifically, we process four types of data \cite{de2018advances}, including fundamental data (e.g., earning reports), market data (e.g., OHLCV data), analytics (e.g., news sentiment), and alternative data (e.g., social media data, ESG data). In addition, we incorporate NLP features for financial sentiment analysis. We build market environments using these features, by following OpenAI gym-style APIs~\cite{brockman2016openai}.



\subsection{Data Layer for Processing and Engineering Highly Unstructured Financial Big Data}

Following the training data development pipeline in data-centric AI~\cite{zha2023data}, we develop a data layer using DataOps practices~\cite{ereth2018dataops}, shown in Fig. \ref{fig:datalayer}. We establish a standard pipeline for financial data engineering, which processes data from different sources into a unified market environment. The pipeline involves a data accessing module to collect data, as well as data cleaning and feature engineering modules to prepare data and transform it into a form that is appropriate for FinRL model training.

\begin{table*}[t] 
    \centering
    \renewcommand{\arraystretch}{1.35}
\caption{Supported data sources. OHLCV means open, high, low, and close prices; volume data.}
\resizebox{1\textwidth}{!}{
\begin{tabular}{c c c c c}
\hline
\textbf{Data Source} & \textbf{Type} & \textbf{Max Frequency} & \textbf{Raw Data} & \textbf{Preprocessed Data}\\
\hline
    Alpaca &  US Stocks, ETFs &  1 min &  OHLCV &  Prices, indicators \\
    Baostock &  CN Securities &  5 min &  OHLCV &  Prices, indicators \\
    Binance &  Cryptocurrency &  1 s &  OHLCV &  Prices, indicators \\
    CCXT &  Cryptocurrency &  1 min  &  OHLCV &  Prices, indicators \\
    IEXCloud &  NMS US securities & 1 day  & OHLCV &  Prices, indicators \\
    JoinQuant &  CN Securities &  1 min  &  OHLCV &  Prices, indicators \\
    QuantConnect &  US Securities &  1 s &  OHLCV &  Prices, indicators \\
    RiceQuant &  CN Securities &  1 ms  &  OHLCV &  Prices, indicators \\
    Tushare &  CN Securities & 1 min  &  OHLCV &  Prices, indicators \\
    WRDS &  US Securities &  1 ms  &  Intraday Trades & Prices, indicators \\
    YahooFinance &  US Securities & 1 min  &  OHLCV  &  Prices, indicators \\
    AkShare &  CN Securities & 1 day  &  OHLCV &  Prices, indicators \\
    findatapy &  CN Securities & 1 day  &  OHLCV &  Prices, indicators \\
    pandas\_datareader &  US Securities &  1 day &  OHLCV & Prices, indicators \\
    pandas-finance &  US Securities &  1 day  &  OHLCV  & Prices, indicators \\
    ystockquote &  US Securities &  1 day  &  OHLCV & Prices, indicators \\
    Marketstack & 50+ countries &  1 day  &  OHLCV & Prices, indicators \\
    finnhub & US Stocks, currencies, crypto &   1 day &  OHLCV  & Prices, indicators \\
    Financial Modeling prep & US stocks, currencies, crypto &  1 min &  OHLCV  & Prices, indicators \\
    EOD Historical Data &  US stocks, and ETFs &  1 day  &  OHLCV  & Prices, indicators \\
    Alpha Vantage & Stock, ETF, forex, crypto, technical indicators &  1 min &  OHLCV  & Prices, indicators \\
    Tiingo & Stocks, crypto &  1 day  &  OHLCV  & Prices, indicators \\
    Quandl & 250+ sources &  1 day  &  OHLCV  & Prices, indicators \\
    Polygon &  US Securities &  1 day  &  OHLCV  & Prices, indicators \\
    fixer &  Exchange rate &  1 day &  Exchange rate & Exchange rate, indicators \\
    Exchangerates &  Exchange rate &  1 day  &  Exchange rate & Exchange rate, indicators \\
    Fixer &  Exchange rate &  1 day  &  Exchange rate & Exchange rate, indicators \\
    currencylayer &  Exchange rate & 1 day  &  Exchange rate & Exchange rate, indicators \\
    currencyapi &  Exchange rate & 1 day &  Exchange rate & Exchange rate, indicators \\
    Open Exchange Rates &  Exchange rate &  1 day  &  Exchange rate & Exchange rate, indicators \\
    XE &  Exchange rate &  1 day  &  Exchange rate & Exchange rate, indicators \\
    Xignite &  Exchange rate &  1 day  &  Exchange rate & Exchange rate, indicators \\
\hline
\end{tabular}}\label{tab:data_sources}
\end{table*}

\subsubsection{Data Accessing}

Users can connect data APIs of different market platforms in Table \ref{tab:data_sources} via our common interfaces. Users can access data agilely by specifying the start date, end date, stock list, time interval, and other parameters. FinRL-Meta has supported more than $30$ data sources, covering stocks, cryptocurrencies, ETFs, forex, etc. 

\subsubsection{Data Cleaning}

Raw data retrieved from different data sources are usually of various formats and with erroneous or missing data to different extents. It makes data cleaning highly time-consuming. With a data processor, we automate the data-cleaning process. In addition, we use stock ticker names and data frequency as unique identifiers to merge all types of data into a unified data table.

\subsubsection{Feature Engineering}

Feature engineering is an important step in training data development~\cite{zha2023data}. Deep learning-based feature engineering has great potential to automate the design of technical indicators \cite{xiao2020feature, nargesian2017Feature}. FinRL-Meta aggregates effective features to help improve model predictive performance. FinRL-Meta currently supports five types of features:
\begin{itemize} [leftmargin=*]

\item \textbf{Market features}:
 Open-high-low-close price and volume data are the typical market data we can directly get from querying the data API. They have various data frequencies, such as daily prices from YahooFinance, and TAQ (Millisecond Trade and Quote) from WRDS. In addition, we automate the calculation of technical indicators based on OHLCV data by connecting the Stockstats\footnote{Github repo: \url{https://github.com/jealous/stockstats}} or TA-lib library\footnote{Github repo: \url{https://github.com/mrjbq7/ta-lib}} in our data processor, such as Moving Average Convergence Divergence (MACD), Average Directional Index (ADX), Commodity Channel Index (CCI), etc.

\item \textbf{Fundamental features}:
Fundamental features are processed based on the earnings data in SEC filings queried from WRDS. The data frequency is low, typically quarterly, e.g., four data points in a year. To avoid information leakage, we use a two-month lag beyond the standard quarter end date, e.g., Apple released its earnings report on 2022/07/28 for the third quarter (2022/06/25) of the year 2022. Thus for the quarter between 04/01 and 06/30, our trade date is adjusted to 09/01 (same method for the other three quarters). We also provide functions in our data processor for calculating financial ratios based on earnings data such as earnings per share (EPS), return on asset (ROA), price to earnings (P/E) ratio, net profit margin, quick ratio, etc.

\item \textbf{Analytics features}:
We provide news sentiment for analytics features. First, we get the news headline and content from WRDS \cite{xinyi_2019}. Next, we use NLTK.Vader\footnote{Github repo: \url{https://github.com/nltk/nltk}} to calculate sentiment based on the sentiment compound score of a span of text by normalizing the emotion intensity (positive, negative, neutral) of each word. For the time alignment with market data, we use the exact enter time, i.e., when the news enters the database and becomes available, to match the trade time. For example, if the trade time is every ten minutes, we collect the previous ten minutes' news based on the enter time; if no news is detected, then we fill the sentiment with 0.
\item \textbf{Alternative features}:
Alternative features are useful but hard to obtain from different data sources \cite{de2018advances}, such as ESG data, social media data, Google trend searches, etc. ESG (Environmental, social, governance) data are widely used to measure the sustainability and societal impacts of investment. The ESG data we provide is from the Microsoft Academic Graph database, which is an open-resource database with records of scholarly publications. We have functions in our data processor to extract AI publication and patent data, such as paper citations, publication counts, patent counts, etc. We believe these features reflect companies' research and development capacity for AI technologies \cite{fang2019practical,chen2020quantifying}. It is a good reflection of ESG research commitment.
\item \textbf{Natural language processing (NLP) features}: 
We employ NLP methods to extract patterns from many sources such as Twitter, Weibo, Google Trends, and Sina finance. NLP \cite{xing2018natural} greatly improves the efficiency of data processing, and reduces the labor of reading texts, websites, videos, and so on. NLP in finance can provide meaningful insights, e.g., sentiment analysis and question-answering like ChatGPT \cite{ouyang2022training}, and GPT-4 \cite{GPT4} when making decisions. NLP features are extracted from a large amount of raw data, and reflect the states, predictions, and emotions of traders, governments, financial institutions, etc. We list several NLP features here: number of comments, number of replies, number of praise/dispraise, and number of optimism/pessimism. 

\end{itemize}

Apart from the above default features, users can quickly add customized features using open-source libraries or add user-defined features. New features can be added in two ways: 1) Write a user-defined feature extraction function. The returned features are added to a feature array. 2) Store the features in a file, and put it in a default folder. Then, an agent can read these features from the file.

\subsection{Financial Sentiment Analysis}
\label{sect:NLP_sentiment}

In addition to the features collected directly from the market data, we also perform sentiment analysis on other data sources, and the output sentiment score serves as another feature. FinRL-Meta investigates the potential extension of sentiment analysis to the financial market context and assesses its impact on automated trading. In addition to market data such as price and volume, NLP features can provide complementary information. The use of market data alone is inadequate in capturing unexpected market events, news, and company announcements, leading to a diminished capacity of trading strategies to respond to unpredictable stock price fluctuations. The incorporation of NLP features, specifically  news, social media, company announcement, and trends sentiments, can help investors analyze market trends and facilitate the examination of the interplay between textual data and stock prices.

Previous studies have extensively investigated the use of NLP techniques and sentiment analysis in stock price prediction and automated trading strategies. In a review study, \cite{xing2018natural} summarized the NLFF (Natural Language Processing for Financial Forecasting) methodologies and their applications in related work. They organized and categorized these techniques into two primary groups: 1) Lexicon-based, which utilizes pre-trained financial sentiment orientation dictionaries \cite{loughran2011liability} to label word segments with sentiment scores; and 2) Automatic Labeling, which employs label propagation frameworks to automatically construct lexicons for the financial domain using seed words \cite{hamilton2016inducing,tai2013automatic}. The lexicon-based approach has the advantage of analyzing texts at the word or sentence level by labeling groups of seed word segments with positive, negative, or neutral sentiments and assigning polarity scores to individual words to represent sentiment strengths. 

Our observation of financial news revealed that the majority of sentiments are expressed through a small group of signaling words related to trading actions, such as `rise' and `drop'. To address this characteristic, we recognized the need for word-level adjustments of sentiment weights for each word segment. As a result, we decided to use the lexicon-based approach in our study. However, we acknowledge that this approach has two major drawbacks: 1) words can have multiple meanings and sentiment strengths in different contexts, and 2) the meaning and sense of a word that is common in one domain, such as e-commerce, may not be common in finance. To overcome these challenges, we explore a specialized framework of lexicon-based sentiment analysis.

In FinRL-Meta, we propose a lexicon-based sentiment analysis framework that is specifically designed for NLP features. To achieve this, we created a customized sentiment dictionary tailored to the characteristics of the financial sector. Our aim is to improve the accuracy of sentiment classification beyond what has been achieved by previous NLP sentiment models by optimizing and extending the existing sentiment analysis framework while adapting it to the textual features of financial news.

\begin{figure}
\centering
\includegraphics[scale=0.4]{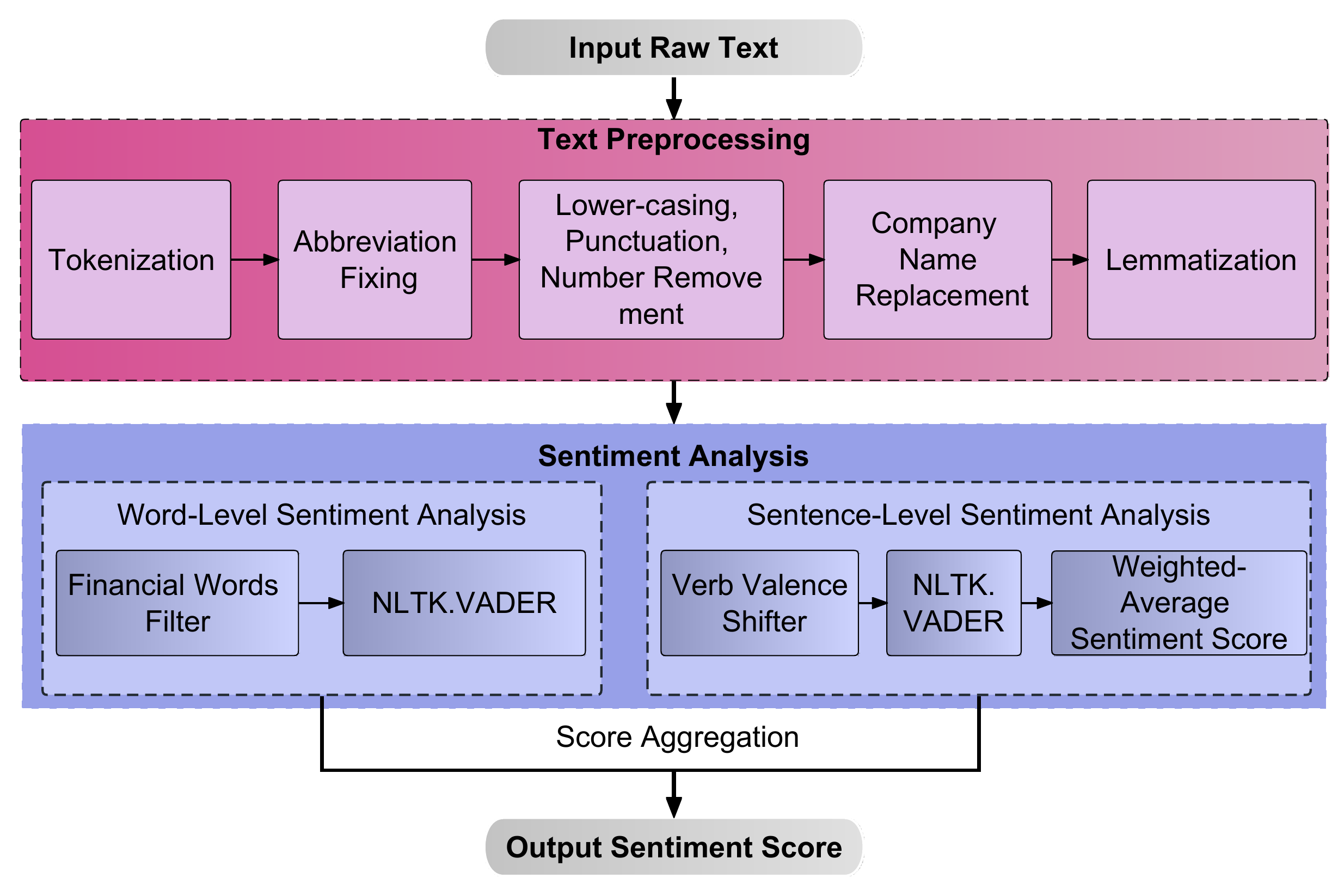}
\caption{Lexicon-based sentiment analysis framework.}
\label{fig:nlp_data}
\end{figure}

\subsubsection{Lexicon-based Sentiment Analysis Framework}
Fig. \ref{fig:nlp_data} illustrates the structure of our lexicon-based sentiment analysis framework, which comprises two main stages: text preprocessing and multi-level sentiment analysis. The text pre-processing stage involves five specific procedures aimed at transforming unstructured news text into cleaned, structured text vectors suitable for sentiment analysis.

\begin{itemize}[leftmargin=*]
\item 1) In the tokenization stage, the raw texts are tokenized to split them into sentence and word segments, which are then respectively parsed into word-level and sentence-level sentiment analysis functions. 
\item 2) Abbreviation Fixing is performed to restore abbreviations containing valuable sentiment signaling words to their complete expressions, enabling the sentiment analysis algorithm to capture all sentiment signaling words.
\item 3) Lower-casing, Punctuation, and Number Removement are conducted to normalize word segments into their lower-cased format and remove unnecessary punctuation and numbers that contain no sentiment values. 
\item 4) Company Name Replacement is carried out to remove company names that contain sentiment-sensitive terms, and replace them with a term containing no sentiment sensitivity, such as ``Company Target'' or ``Company Random''.
\item 5) Lemmatization is employed to group together the inflected forms of a word so they can be analyzed as a single item, identified by the word's lemma. Unlike stemming which reduces all terms to their word stem, lemmatization restores each word to its lemma form based on its part of speech tagging. For instance, the word ``developed'' is
tagged as VERB and restored to the lemma ``develop''. The word ``development'' is tagged
as NOUN and restored to ``development''. Since different inflected forms of a word demonstrate distinctive sentiment polarities, the lemmatization method is used instead of stemming.
\end{itemize}

The sentiment analysis stage of text data involves transforming word and sentence vectors obtained during text preprocessing into corresponding sentiment scores, which are then aggregated to obtain the overall sentiment score. This stage employs distinct methodologies for word and sentence vectors. 

\begin{itemize}[leftmargin=*]
\item 1) For the word vectors, to filter out irrelevant word segments and concentrate on finance-specific sentiment signaling words, we utilized the LoughranMcDonald MasterDictionary \cite{xing2018natural} with over 80,000 core financial terms as the base financial corpus. We then computed word-level sentiment scores using the VADER (Valence Aware Dictionary for Sentiment Reasoning) model in NLTK \cite{hutto2014vader}. 
\item 2) For sentence vectors, we customized and adjusted the intensities of signaling words based on sentence logic and semantics using the Adverb Valence Shifter mechanism in the VADER package \cite{hutto2014vader}. To incorporate sentiment-strengthening or weakening terms like ``hardly'' and ``significantly'', we also implemented the Verb Valence Shifter mechanism which identified the verb and noun words as the sentence separator for each input sentence. After adjusting the internal sentiment weights of each sentence, we devised rules to assign weights to different sentences and aggregate the sentence vectors to obtain sentence-level sentiment scores. 
\end{itemize}

\subsubsection{Financial Sentiment Dictionary Construction}
The sentiment dictionary is the foundation for lexicon-based sentiment analysis, containing information about the emotions or polarity expressed by words, phrases, or concepts. As noted in the previous section, one of the major obstacles associated with the lexical annotation approach is that the sentiment intensity and polarity of a given word often vary across different domains. For example, in e-commerce, the term ``bull'' usually refers to an animal species, while in the finance domain, it signifies an inclination for a specific market, security, or industry to rise.

\begin{figure}
\centering
\includegraphics[scale=0.4]{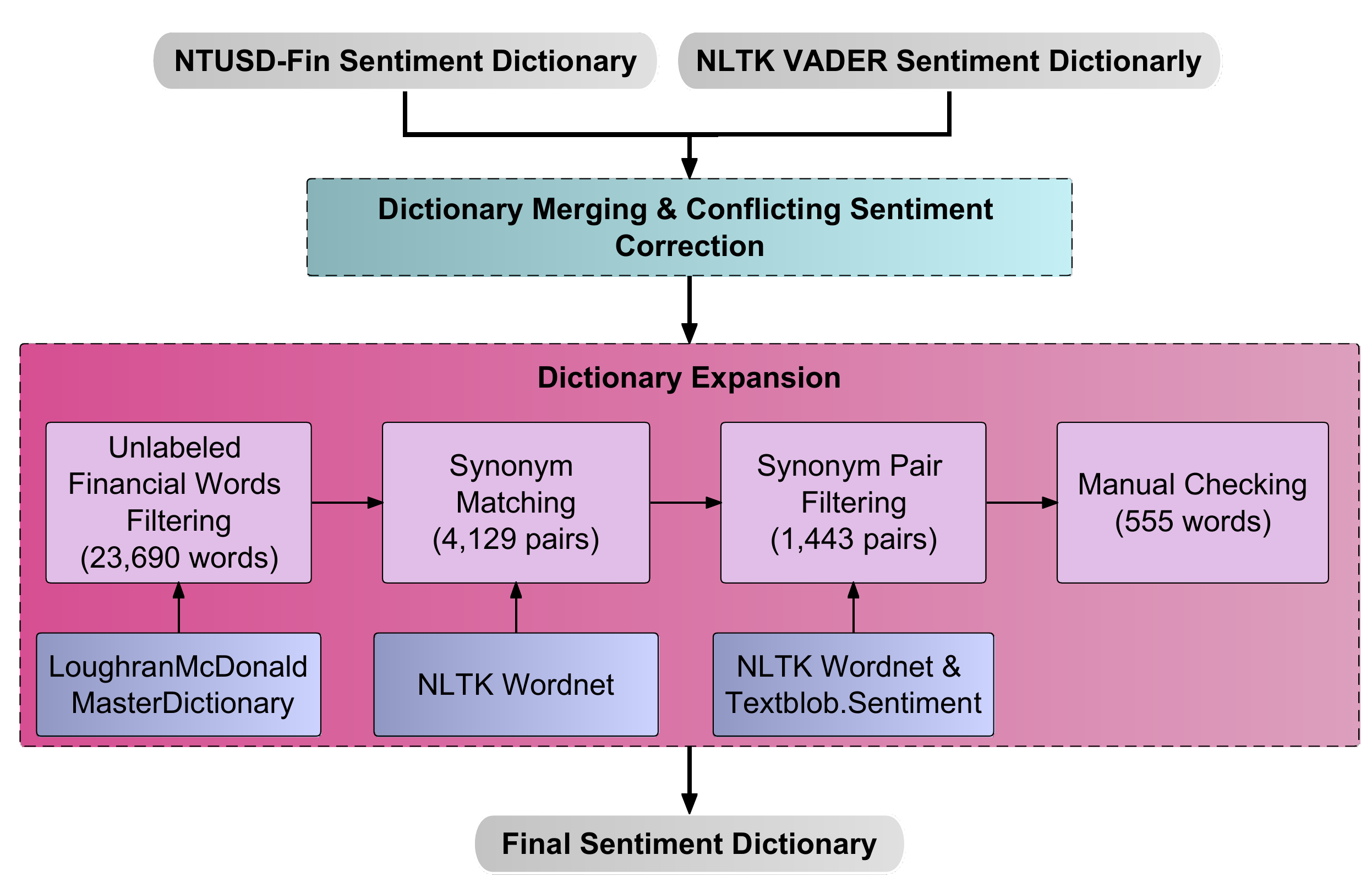}
\caption{Financial sentiment dictionary.}
\label{fig:nlp_dict}
\end{figure}

We create a financial sentiment dictionary by customizing and expanding two existing sentiment dictionaries, as shown in Fig. \ref{fig:nlp_dict}. The construction process involved two stages: dictionary merge and dictionary expansion. During the first stage, we merged the NTUSD-Fin Sentiment Dictionary with the built-in dictionary of the NLTK VADER package. The NTUSD-Fin dictionary \cite{chen2018ntusd} is a sentiment dictionary specifically designed for finance, comprising labels for 8,331 words, 112 hashtags, and 115 emojis, and providing multiple scoring techniques, such as frequency, CFIDF, chi-squared value, market sentiment score, and word vector for tokens. The VADER sentiment dictionary \cite{hutto2014vader}, created based on social media texts, includes 7,502 labeled words. Throughout the merging process, we detected and manually fixed 479 contradictions in the semantic polarities of words between the two dictionaries.

We also observe that certain financial terms with sentiment values were not accurately captured by the merged sentiment corpus, such as ``bullish'', ``bearish'', ``outperform'', ``outgrowth'', ``high'', and ``rise''. Consequently, in the second stage of the construction process, the dictionary was expanded to label more financial words with sentiment polarity and intensities through a synonym-matching mechanism. This process involved four main steps.

\begin{itemize}[leftmargin=*]
\item 1) Unlabeled Financial Words Filtering: the LoughranMcDonald MasterDictionary \cite{xing2018natural}, which contains over 80,000 core financial words, was utilized as the base financial corpus to filter out relevant financial words without attached sentiment polarities, resulting in 23,690 words being filtered out.
\item 2) Synonym Matching: to assign sentiment polarity and intensity scores to the unlabeled words, the NLTK Wordnet was used to match each word with its labeled synonym. The WordNet  \cite{miller1998wordnet} is an extensive lexical database of English words that groups nouns, verbs, adjectives, and adverbs into sets of cognitive synonyms called ``synsets'', each expressing a distinct concept. In this step, the 23,690 unlabeled words were matched with their synonym lists recommended by the WordNet Synset database, and word-synonym pairs with labeled synonyms were selected. This resulted in 4,129 word-synonym pairs being filtered out.
\item 3) Synonym Pair Filtering: an algorithm was developed to further filter out words with the highest sentiment values from the remaining 4,129 word-synonym pairs. This algorithm utilized the TextBlob \cite{loria2018textblob} subjectivity indicator and synonym path similarity indicator to identify the most sentiment-rich word-synonym pairs. The algorithm iterated through all words, checking their related synonyms and selecting the synonym with the highest similarity score in the SentiWordNet synonym network. After computing the subjectivity of each word, those with a subjectivity value of less than 0.2 were excluded. This step resulted in 1,443 word-synonym pairs being filtered out.

\item 4) Manual Checking: each word-synonym-sentiment pair was manually checked, along with its actual usage in the real news context. If the path similarity between the word and its synonym was less than 0.5, its matched sentiment score was adjusted based on its meaning in the context. After this manual checking step, 555 words were selected as the final extended semantic dictionary.
\end{itemize}

To evaluate sentiment analysis accuracy involving both sentiment polarities and strengths,
we conduct an experiment based on the news headlines corpus used in SemEval-2007 dataset \cite{strapparava2007semeval}, which includes 1000 news headlines manually annotated with sentiment scores on six emotion categories: Anger, Disgust, Fear, Joy, Sadness, and Surprise. The sentiment scores range from -100 (highly negative) to 100 (highly positive), with 0 indicating neutral valence. Of the 1000 headlines, 468 have a positive valence, 526 have a negative valence, and 6 have a neutral valence. For this experiment, only headlines with clear positive emotions (sentiment valence of 50 to 100) and negative emotions (sentiment valence of -100 to -50) were selected, resulting in a dataset of 410 headlines, consisting of 155 positive and 255 negative headlines.

The evaluation results of the experiment are presented in Table \ref{event:sentiment_table}. Two metrics were used to assess the accuracy of the sentiment analysis approaches Polarity Accuracy, which measures the proportion of predicted polarity (negative, neutral, and positive) that matches the labeled polarity, and Valence Correlation, which measures the correlation between predicted valence scores (ranging from -1 to 1) and labeled valence scores (ranging from -100 to 100), indicating the accuracy of the prediction of news sentiment intensities. As shown in the table, our sentiment analysis approach outperforms the baseline approach in terms of both Polarity Accuracy and Valence Correlation.

\begin{table*}
\begin{center}
\begin{tabular}{|p{4cm}|p{3.2cm}|p{3.5cm}|}
\hline
\textbf{Approach} &\textbf{Polarity Accuracy} &\textbf{Valence Correlation}\\
\hline
NLTK VADER (Baseline)
& 0.60
&0.66\\
\hline
With Verb Valence Shifter
&  0.62
&0.67\\
\hline
With Financial sentiment dictionary (Our Approach)
&  0.74
&0.70\\
\hline
\end{tabular}
\end{center}
\caption{Evaluation Result of Financial Sentiment Analysis.}
\label{event:sentiment_table} 
\end{table*}

\subsection{Environment Layer for Dynamic Market Environments}
\label{sect:env_layer}

FinRL-Meta follows the OpenAI gym-style \cite{brockman2016openai} to create market environments using the cleaned data from the data layer. It provides hundreds of environments with a common interface. Users can build their environments using FinRL-Meta's interfaces, share their results and compare a strategy's trading performance. Following the gym-style \cite{brockman2016openai}, each environment has three functions as follows:
\begin{itemize}[leftmargin=*]
    \item \texttt{reset()} function resets the environment back to the initial state $s_0$
    \item \texttt{step()} function takes an action $a_t$ from the agent and updates state from $s_t$ to $s_{t+1}$.
    \item \texttt{reward()} function computes the reward value transforming from $s_t$ to $s_{t+1}$ by action $a_t$.
\end{itemize}
Detailed descriptions can be found in \cite{yang2020deep}\cite{gort2022deep}.

We plan to add more environments for users' convenience. For example, we are actively  building market simulators using limit-order-book data (refer to Appx. \ref{subsec:market_simu}), where we simulate the market from the playback of historical limit-order-book-level data and an order matching mechanism. We foresee the flexibility and potential of using a Hidden Markov Model (HMM) \cite{mamon2007hidden}  or a generative adversarial net (GAN) \cite{goodfellow2014generative} to generate market scenarios \cite{coletta2021towards}.

\textbf{Incorporating trading constraints to model market frictions}:
To better simulate real-world markets, we incorporate common market frictions (e.g., transaction costs and investor risk aversion) and portfolio restrictions (e.g., non-negative balance). 
\begin{itemize}[leftmargin=*]
\item \textbf{Flexible account settings}: Users can choose whether to allow buying on margin or short-selling.
\item \textbf{Transaction cost}: We incorporate the transaction cost to reflect market friction, e.g., $0.1\%$ of each buy or sell trade.
\item \textbf{Risk-control for market crash}: In FinRL \cite{liu2020finrl,liu2021finrl}, a turbulence index \cite{kritzman2010skulls} is used to control risk during market crash situations. However, calculating the turbulence index is time-consuming. It may take minutes, which is not suitable for paper trading and live trading. We replace the financial turbulence index  with the volatility index (VIX) \cite{whaley2009understanding} that can be accessed immediately.
\end{itemize}

\textbf{Multiprocessing training via vectorized environments}:
We utilize GPUs for multiprocessing training, namely, the vectorized environments technique of Isaac Gym \cite{makoviychuk2021isaac}, which significantly accelerates the training process. In each CUDA core, a trading agent interacts with a market environment to produce transitions in the form of $\{$state, action, reward, next state$\}$. Then, all the transitions are stored in a replay buffer and later are used to update a learner. By adopting this technique in our market simulator, we successfully achieve the multiprocessing simulation of hundreds of market environments to improve the performance of DRL trading agents on large datasets.

\section{Homegrown Examples and Tutorials} \label{benchmarks}

We provide several homegrown examples and a dozen of tutorials, which serve as stepping stones for newcomers. We reproduce popular papers as benchmarks for follow-up research.

\subsection{Performance Metrics and Baseline Methods}
\label{sec:performance_metrics}

We provide the following metrics to measure the trading performance: 
\begin{itemize}[leftmargin=*]
  \item \textbf{Cumulative return} $R = \frac{v - v_0}{v_0}$, where $v$ is the final portfolio value, and $v_0$ is the original capital.
  \item \textbf{Annualized return} $r = (1+R)^\frac{365}{t}-1$, where $t$ is the number of trading days.
  \item \textbf{Annualized volatility} ${\sigma}_a = \sqrt{\frac{\sum_{i=1}^{n}{(r_i-\bar{r})^2}}{n-1}}$, where $r_i$ is the annualized return in year $i$, $\bar{r}$ is the average annualized return, and $n$ is the number of years.
  \item \textbf{Sharpe ratio} \cite{Sharpe} 
$S_T = \frac{\text{mean}(R_t) - r_f}{\text{std}(R_t)}$, where $R_t = \frac{v_t - v_{t-1}}{v_{t-1}}$, $r_f$ is the risk-free rate, and $t=1,...,T$.
  \item \textbf{Max. drawdown}: The maximal percentage loss in portfolio value.
\end{itemize}

The following baseline trading strategies are provided for comparison:
\begin{itemize}[leftmargin=*]
    \item \textbf{Passive trading strategy} \cite{malkiel2003passive} is a well-known long-term strategy. The investors just buy and hold selected stocks or indexes without further activities.
    \item \textbf{Mean-variance and min-variance strategy} \cite{ang2012mean} are two widely used strategies that look for a balance between risks and profits. They select a diversified portfolio in order to achieve higher profits at a lower risk.
    \item \textbf{Equally weighted strategy} is a portfolio allocation strategy that gives equal weights to different assets, avoiding allocating overly high weights on particular stocks. 
\end{itemize}

\subsection{Homegrown Examples}

For educational purposes, we provide Jupyter/Python as tutorials\footnote{https://github.com/AI4Finance-Foundation/FinRL-Tutorials} to help newcomers get familiar with the whole pipeline.

In Section \ref{sec_RLOps}, we have described the RLOps paradigm. Here we will demonstrate how to use RLOps in practice by reproducing several prior papers. We have reproduced experiments in several papers as benchmarks. Users can study our codes for research purposes or use them as stepping stones for deploying trading strategies in live markets. In this subsection, we describe several home-grown examples in detail, in a sequence of simple to advanced. Users could choose the one to learn and run based on their proficiency. 

\begin{figure}
\centering
\includegraphics[scale=0.07]{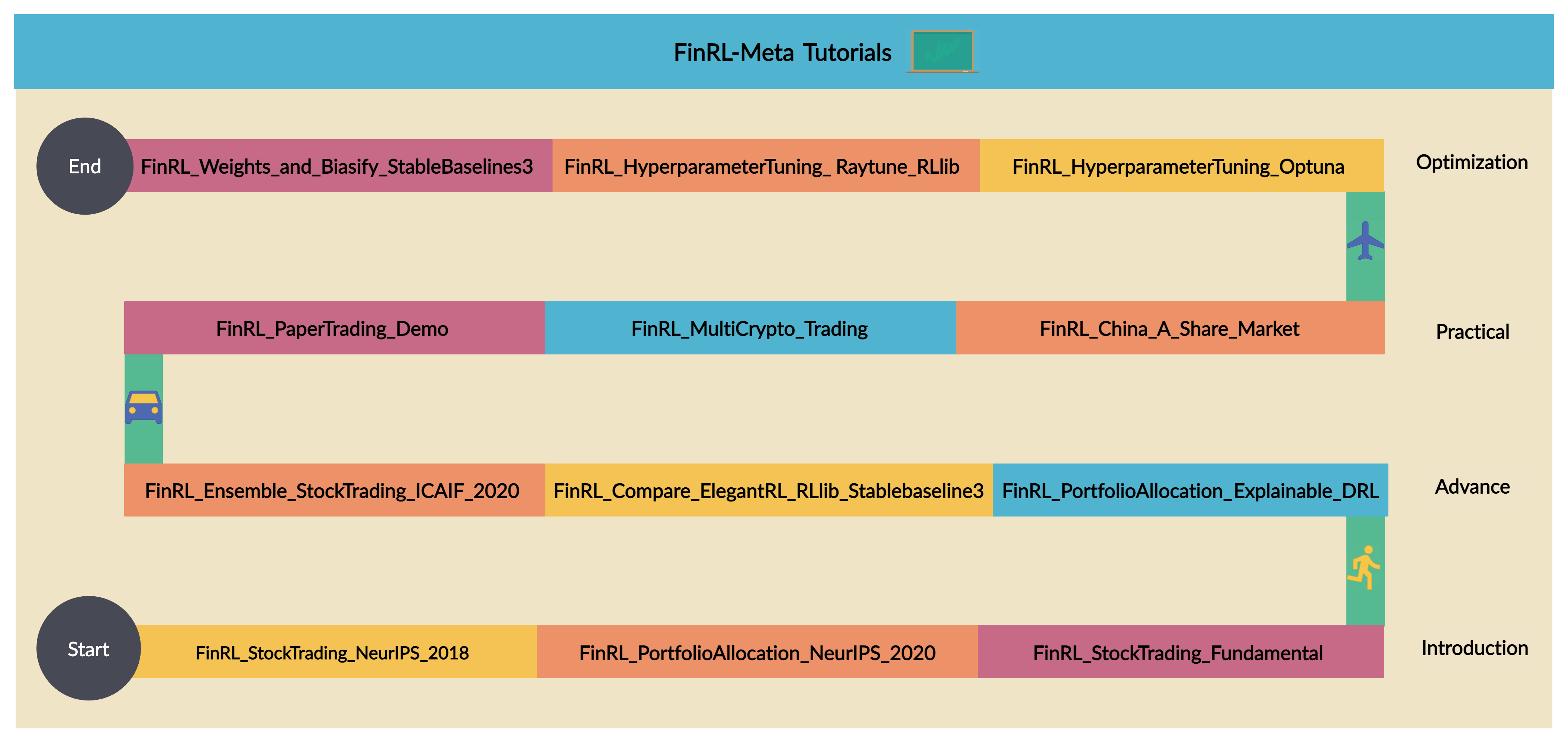}
\caption{Demos of FinRL-Meta, organized in a curriculum structure.}
\label{fig:tutorials}
\end{figure}
\begin{figure}
\centering
\includegraphics[scale=0.3]{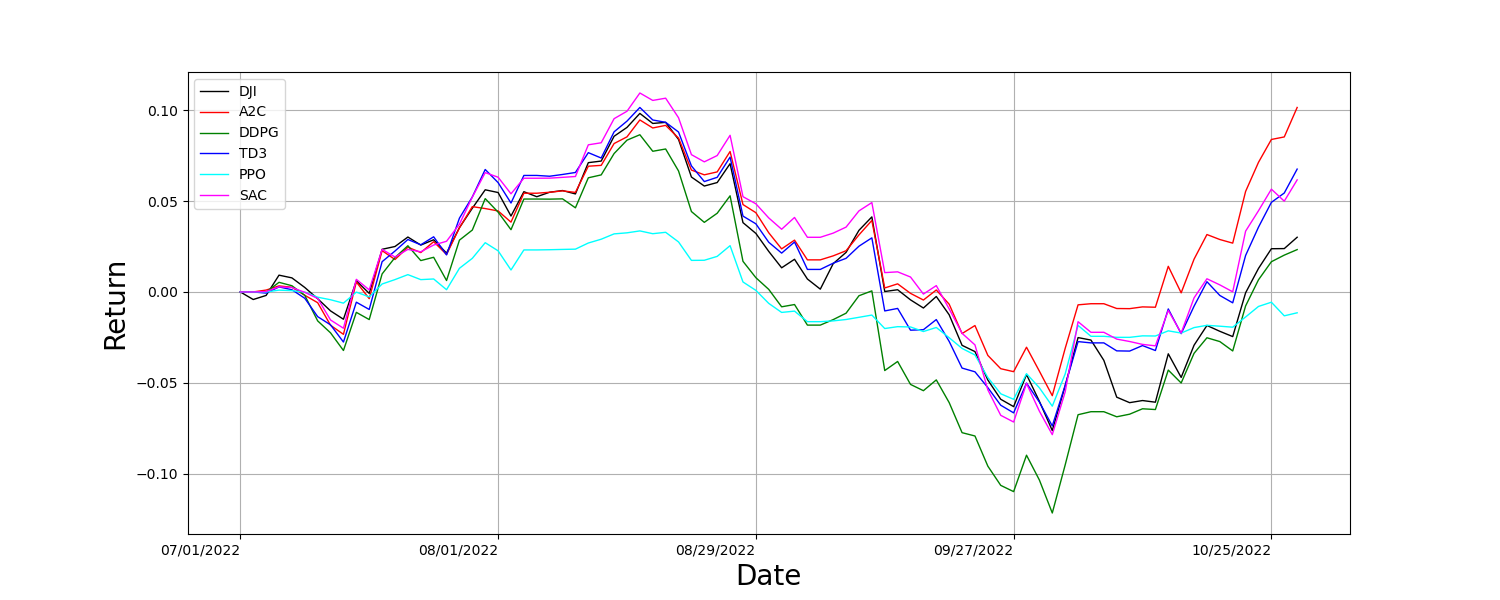}
\caption{Reproducing the stock trading of \cite{liu2018practical}.}
\label{fig:stock_trading_performance}
\end{figure}

\subsubsection{Stock Trading Task}
The stock trading task is one of the most classical problems in FinRL-Meta. The goal is to train a DRL agent to decide the time, ticker, and amount to buy and sell on the stock market.

First, users can use the APIs provided by FinRL-Meta to fetch historical OHLCV (open, high, low, close prices, and volume) data of the stocks from data platforms like Yahoo Finance. After data cleaning and feature engineering, which checks the error and missing data, and then technical indicators are added.

Second, FinRL-Meta will split the processed data into training and testing, and construct a gym-style market environment for each, with the state-action-reward specified in Section \ref{subsec:modeling_finrl}. Then, we are able to choose algorithms from any one of the DRL libraries and train the agents of these algorithms in the market environment.

Lastly, after we obtain the trained agents, we backtest the agents on the environment with testing data. During backtesting, the performance of the agents will be compared with the performance of several baselines, such as DJIA index, mean-variance, and equally weighted.

In FinRL-Meta's demo, after fetching from Yahoo! Finance, we use the data from 01/01/2011 to 07/01/2021 for training and the data from 07/01/2021 to 11/01/2022 for trading. The technical indicators in the state space include the following, Moving Average Convergence Divergence (MACD), Relative Strength Index (RSI), Commodity Channel Index (CCI), Average Directional Index (ADX), etc. All these data will be sealed in a gym-style environment.

During the training phase, we utilized dynamic datasets with a rolling window of 22 trading days and trained five different deep reinforcement learning (DRL) algorithms (A2C, DDPG, TD3, PPO, and SAC) in the environment. Fig.~\ref{fig:stock_trading_performance} displays the comparison of their performances with the baseline DJIA index after backtesting all the agents. The best agent was found to be A2C, achieving a return of 0.102 compared to DJI's return of 0.030. This demo provides a detailed walkthrough of how DRL operates in the stock trading task and serves as a benchmark for subsequent works in the field of financial reinforcement learning \cite{liu2018practical}.  This benchmark is beneficial for getting into the field of financial reinforcement learning.

\subsubsection{Trading in Real Time}

\begin{figure}[t]
\centering
\includegraphics[scale=0.45]{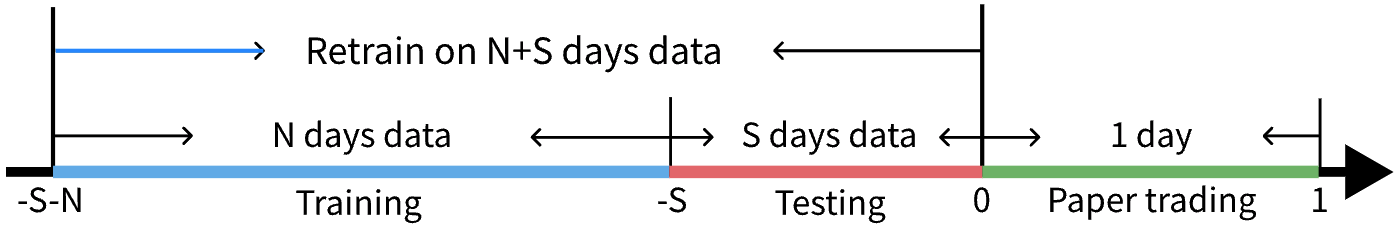}
\caption{Data split for a window that consists of training, testing, and trading.}
\label{fig:rw_split}
\end{figure}

Practical tasks like stock trading and cryptocurrency trading suffer from the false positive issue due to overfitting, where an agent might perform well on testing data but not on real-world markets. Since backtesting has problems of information leakage and overfitting, its results is not persuasive to show the quality of trained models. Thus, we propose to deploy DRL agent on paper trading. 

Fig. \ref{fig:rw_split} shows our ``training-testing-trading'' pipeline. In a window, there are $N$ days' data for training and $S$ days' data for testing. At the end of a window, we perform paper trading for $1$ day. Note that we always retrain the agent using $N+S$ days of training and testing data together. Then, we roll the window forward by $1$ day ahead and perform the above steps for a new window. Paper trading is always carried out for $1$ day. Therefore, $D$ windows correspond to $D$ trading days. 

Alg. \ref{algo_pt} summarizes the pipeline of paper trading. For $D$ trading days from $0$ to $D-1$, we keep doing the following three steps:

\begin{itemize}[leftmargin=*]
    \item \textbf{Step 1)}. Download and process $N$-day data, from day $d-S-N$ to day $d-S-1$. Then build the data into a gym-style environment and train the agent. Then download and process S-day data, from day $d-S$ to day $d-1$. Then build the data into a gym-style environment and validate how the agent performs. According to the agent's performance on the validation environment, adjust hyper-parameters.
    \item \textbf{Step 2)}. Build the training and testing data, totally $N+S$ days from day $d-S-N$ to day $d-1$ (note that there are $390$ data points for each day's minute-level data), into a gym-style environment. Update hyper-parameters to the values chosen from \textbf{Step 1)}. Then retrain the agent on these $ N+S$-day environments.
    \item \textbf{Step 3)}. Deploy the trained agent to the paper trading market and trade from 9:30 am to 4:00 pm.
\end{itemize}

\begin{algorithm}[t]
\caption{Algorithm for stock trading in real time}
\label{algo_pt}
\begin{algorithmic}[1]
    \State Initialize a set of hyper-parameters;
    \For{$d = 0$ to $D-1$}
        \State \# \textbf{Step 1)}. Train an agent
	\State Using data period $[d-S-N, d-S-1]$ to train an agent,
        \State Using data period $[d-S, d-1]$ to test the trained agent and adjust hyper-parameters,
        \State \# \textbf{Step 2)}. Retrain the agent
        \State Using data period $[d-S-N, d-1]$ to retrain the agent with the adjusted hyper-parameters,
        \State \# \textbf{Step 3)}. Perform paper trading for one day
        \State Using the trained agent to trade in the $d$-th day.
    \EndFor
\end{algorithmic}
\end{algorithm}

 \begin{figure}[t]
\centering
\includegraphics[scale=0.65]{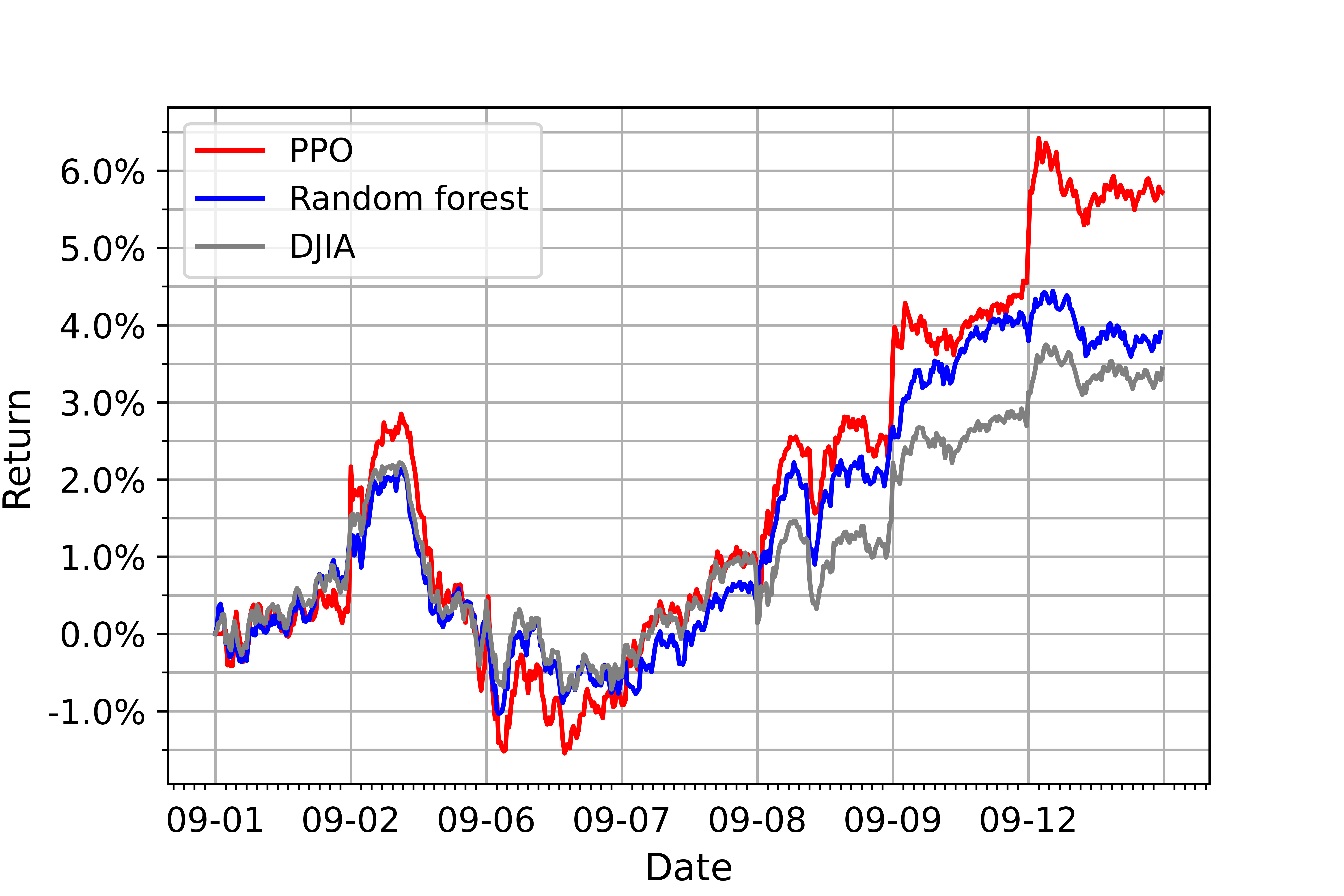}
\caption{Stock trading in real-time: cumulative returns for a conventional ML method (random forest) VS. our trained PPO agent.}
\label{comparing}
\end{figure}

For the experiment, we select Dow Jones 30 stocks as our trading stocks and use minute-level historical market data from $08/24/2022$ to $09/09/2022$. The data are downloaded from Alpaca\footnote{Web page of Alpaca: https://alpaca.markets/}. Then, we use the paper trading APIs provided by Alpaca to do paper trading from $09/01/2022$ to $09/12/2022$. The cumulative return of our PPO method and conventional random forest method is shown in Fig. \ref{comparing}.

\subsubsection{Ensemble Strategy}

Based on the stock trading task, the ensemble method \cite{yang2020deep} combines different agents to obtain an adaptive one, which inherits the best features of agents and performs remarkably well in practice. We consider three component algorithms, Proximal Policy Optimization (PPO), Advantage Actor-Critic (A2C), and Deep Deterministic Policy Gradient (DDPG), which have different strengths and weaknesses. Using a rolling window, an ensemble agent automatically selects the best model for each test period. Again on the 30 constituent stocks of the DJIA index, we use data from 01/01/2010 to 07/01/2022 for training and data from 07/01/2022 to 01/01/2023 for validation and testing through a quarterly rolling window.

\begin{figure}
\centering
\includegraphics[scale=0.3]{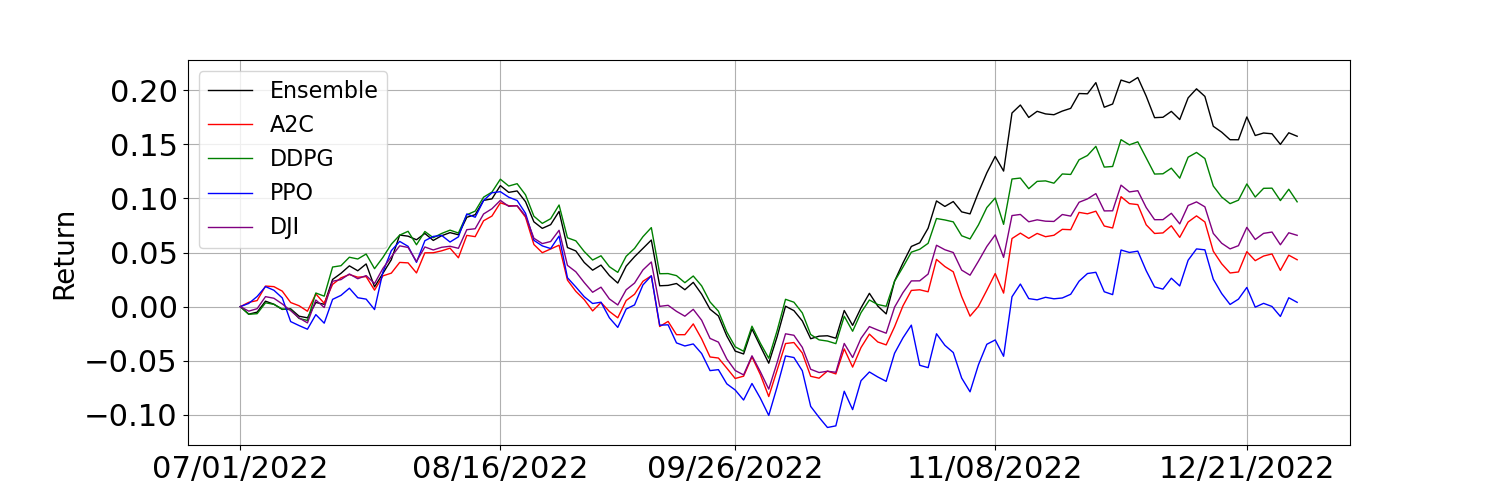}
\caption{Reproducing the ensemble strategy of \cite{yang2020deep}.}
\label{fig:compare_returns_ensemble}
\end{figure}

From Fig.~\ref{fig:compare_returns_ensemble}, we observe that the ensemble agent outperforms other agents. In the experiment, the return of the ensemble strategy is $0.157$, while DJI is $0.068$. The ensemble agent has the highest return, which means it performs the best in profits. This benchmark demonstrates that the ensemble strategy is effective in constructing a more reliable agent based on several components of DRL agents.\looseness=-1

\subsubsection{Podracer on Cloud }

We reproduce cloud solutions of population-based training, e.g., generational evolution \citep{finrl_podracer_2021} and tournament-based evolution \citep{liu2021podracer}.  FinRL-Podracer can easily scale out to $\geq 1000$ GPUs, which features high scalability, elasticity and accessibility by following the cloud-native principle. If GPUs are abundant, users can take advantage of this benchmark to work on high-frequency trading tasks. On an NVIDIA SuperPOD cloud, we conducted extensive experiments on stock trading and found that it substantially outperforms competitors, such as OpenAI and RLlib \cite{finrl_podracer_2021}. Detailed instructions are provided on our website.

\textbf{Benchmarks on cloud}: We provide demos on a cloud platform, Weights \& Biases \footnote{Website: \url{https://wandb.ai/site}}, to demonstrate the training process. We define the hyperparameter sweep, training function, and initialize an agent to train and tune hyperparameters. On the cloud platform Weights \& Biases, users are able to visualize their results and assess  the relative performance via community-wise competitions.

\subsubsection{Curriculum Learning for Generalizable Agents}

Based on FinRL-Meta (a universe of market environments, say $\geq 100$), one is able to construct an environment by sampling data from multiple market datasets, similar to XLand \cite{team2021open}. In this way, one can apply the curriculum learning method \cite{team2021open} to train a generally capable agent for several financial tasks.



\subsubsection{Market Simulator}\label{subsec:market_simu}

\textbf{Gym market environments}: We build all of our environments following OpenAI-gym style \cite{brockman2016openai}. The first reason is that this makes it convenient for plugging in any of the three DRL libraries (Stable Baseline3, RLlib, and ElegantRL). Another reason is that it is user-friendly. Newcomers can learn our environments fast and then build their own task-specific environments efficiently.

\textbf{Synthetic data generation}: We create a market simulator \cite{market_simulator} to simulate the markets from the playback of historical limit-order-book-level data and the order matching mechanism. Currently, the simulator is at the minute level (i.e., one time step = one minute), which is changeable. The state is a stack of market indicators and market snapshots from the last few time steps. The action is to place an order. We support market orders and limit orders. We also provide several wrappers to accept typically discrete or continuous actions. Rewards can be configured by the participants with the aim of generating policies that optimize pre-specified indicators. In our simulator, we take into account the following factors: 1) temporary market impact; 2) order delay. We do not consider the following factors in our simulator: 1) permanent market impact of limit orders; 2) non-resiliency limit order book.

\begin{figure}
\centering
\includegraphics[scale=0.5]{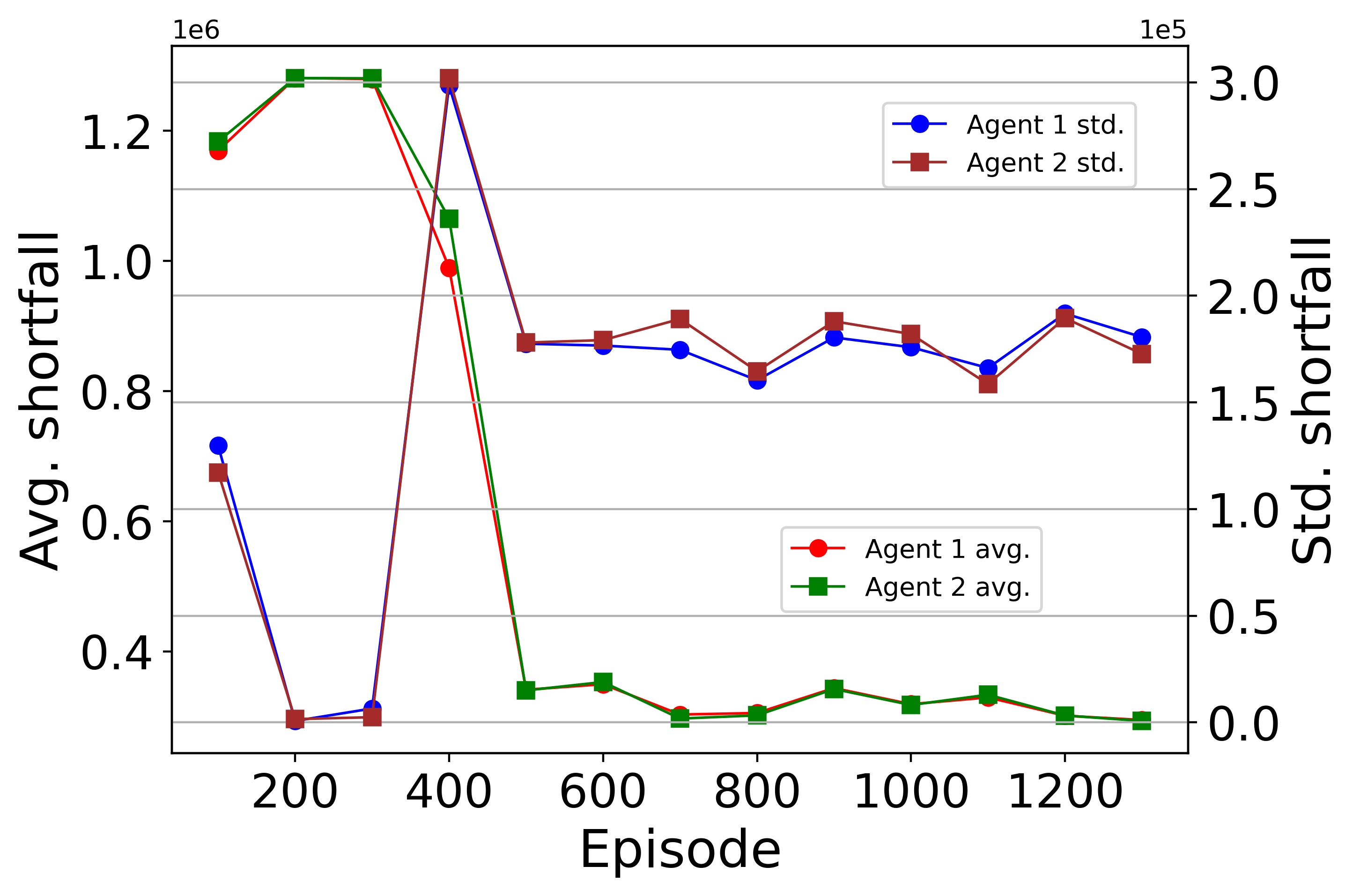}
\caption{Liquidation analysis of \cite{bao2019multiagent}.}
\label{fig:liquid_performance}
\vspace{-4mm}
\end{figure}

\textbf{Liquidation analysis and trade execution:} By reproducing \cite{bao2019multiagent}, we build a simulated environment of stock prices according to the Almgren and Chriss model. Then we implement the multi-agent DRL algorithms for both competing and cooperative liquidation strategies. This benchmark demonstrates the trade execution task using deep reinforcement learning algorithms. When trading, traders want to minimize the expected trading cost, which is also called implementation shortfall. In Fig.~\ref{fig:liquid_performance}, there are two agents, and we observe that the implementation shortfalls decrease during the training process.

\subsection{More FinRL Demos and Tutorials}

\begin{figure}
\centering
\includegraphics[scale=0.3]{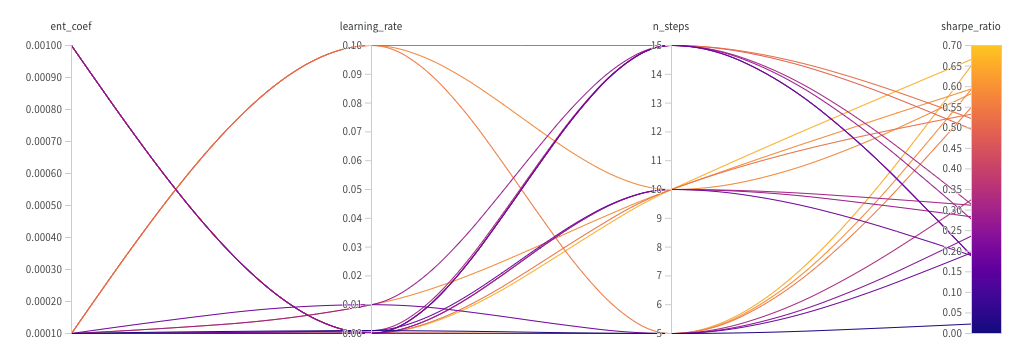}
\caption{An example of hyperparameter tuning.} 
\label{fig:hyperparameter}
\vspace{-2mm}
\end{figure}

\begin{itemize}[leftmargin=*]
    \item \textbf{Portfolio allocation} \cite{liu2020finrl}: We train a DRL agent to perform a portfolio optimization task on a set of stocks.
    \item \textbf{Cryptocurrency trading} \cite{liu2020finrl}: We provide a demo \cite{liu2020finrl} on $10$ popular cryptocurrencies.
    \item \textbf{Paper trading demo}: We provide a demo for paper trading. Users could combine their own strategies or trained agents in paper trading.
    \item \textbf{China A-share demo}: We provide a demo based on the China A-share market data using Tushare.
    \item \textbf{Hyperparameter tuning}: The default hyperparamters may not be the best for our tasks. Reinforcement learning algorithms are sensitive to hyperparamters; therefore, hyperparamter tuning is an important issue. Hyperparamters are tuned based on an objective. We provide several demos for hyperparameter tuning using Optuna \cite{akiba2019optuna} and Ray Tune \cite{liaw2018tune}. Fig.~\ref{fig:hyperparameter} shows an example of hyperparameter tuning in the portfolio allocation task using A2C, which aims to maximize the Sharpe ratio.
    \item \textbf{Robo-advising}: Robo-advising by integrating ChatGPT \cite{ouyang2022training} and GPT-4 \cite{GPT4} will be much more powerful than before. Users are encouraged to develop robo-advisor apps by building upon the demos that we have created using ChatGPT \cite{ouyang2022training} and GPT-4 \cite{GPT4}. By engaging in conversation with ChatGPT using a chain of thought prompt, we may be able to obtain good trading advice. Additionally, we have demonstrated that properly guiding ChatGPT can generate new financial factors, which speeds up the process of creating them manually.
    
\end{itemize}

\section{Conclusion and Future Works}

Following the data-centric AI principles and the DataOps paradigm, in this paper, we developed FinRL-Meta, a data-centric library that provides openly accessible dynamic financial datasets and reproducible benchmarks. FinRL-Meta implements an automated data-curation pipeline tailored for the financial market, which processes data from different sources and integrates them into a unified market environment. FinRL-Meta adheres to the standard OpenAI gym-style APIs, making it readily accessible to experts in both reinforcement learning and finance. In addition, we have developed a wide range of homegrown examples and tutorials to facilitate the usage of the library.  FinRL-Meta will serve as a source of inspiration for both researchers and practitioners who are interested in FinRL.

For future work, FinRL-Meta aims to build a universe of financial market environments, like the XLand environment \cite{team2021open}. To improve the performance for large-scale markets, we are exploiting GPU-based massive parallel simulation such as Isaac Gym \cite{makoviychuk2021isaac} and deploying it into projects such as RL for market simulator. Moreover, it will be interesting to explore the evolutionary perspectives \cite{gupta2021embodied,scholl2021market,finrl_podracer_2021, liu2021podracer} to simulate the markets. We believe that FinRL-Meta will provide insights into complex market phenomena and offer guidance for financial regulations.

\subsection{Explainability}

\begin{figure}
\centering
\includegraphics[scale=0.4]{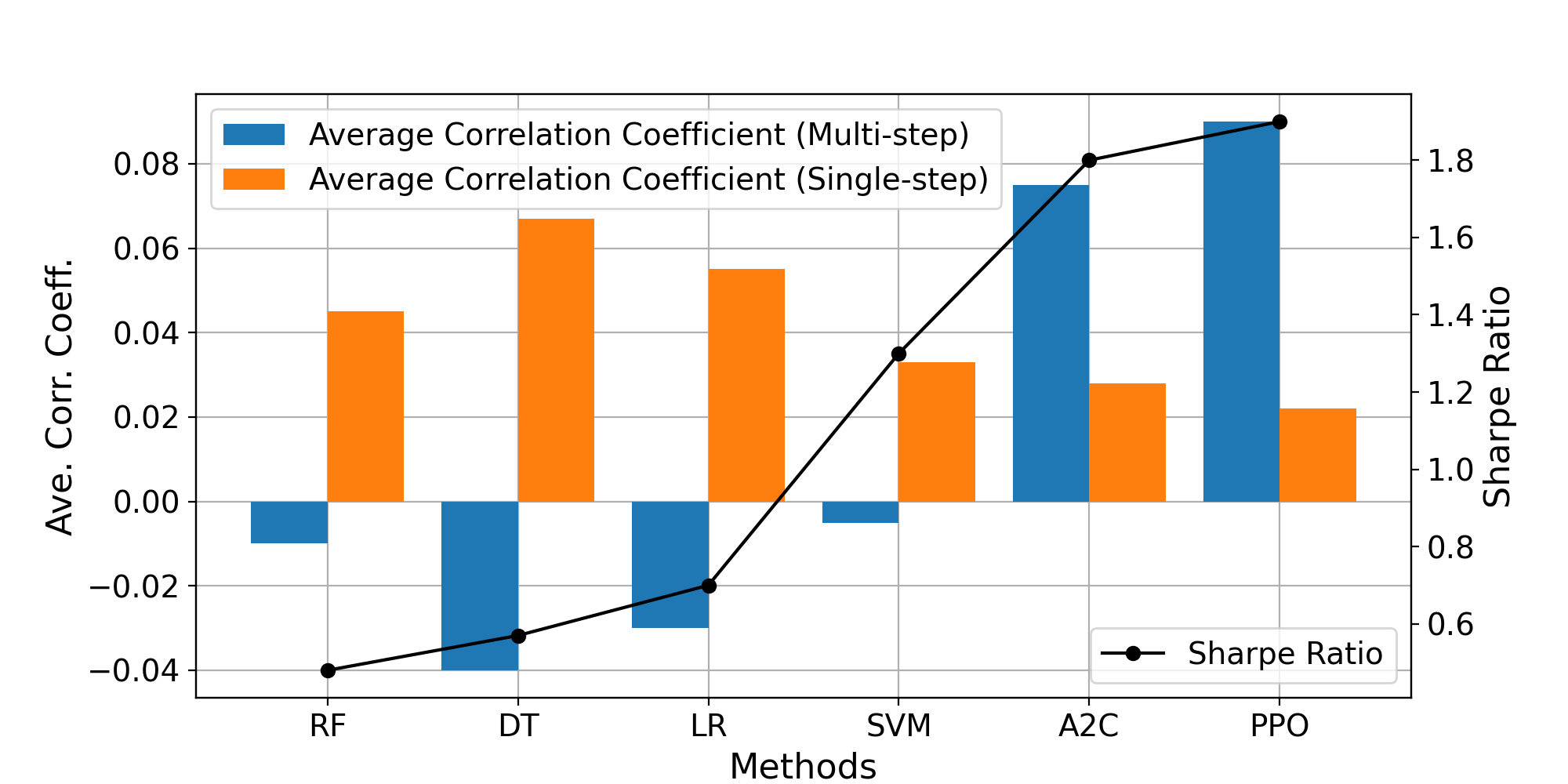}
\caption{Reproducing portfolio management of \cite{guan2021explainable}: Comparison of average correlation coefficient and Sharpe ratio among ML and DRL methods.} 
\label{fig:explainable_CumulativeReturn}
\vspace{-2mm}
\end{figure}

We reproduce \cite{guan2021explainable} that compares the performance of DRL algorithms with machine learning (ML) methods on the multi-step prediction in the portfolio allocation task. We use four technical indicators MACD, RSI, CCI, and ADX as features. Random Forest (RF), Decision Tree Regression (DT), Linear Regression (LR), and Support Vector Machine (SVM) are the ML algorithms in comparison. We use data from Dow Jones 30 constituent stocks to construct the environment. We use data from 04/01/2009 to 03/31/2020 as the training set and data from 04/01/2020 to 05/31/2022 for backtesting. In Fig.~\ref{fig:explainable_CumulativeReturn}, the results show that DRL methods have a higher Sharpe ratio than ML methods. Also, DRL methods' average correlation coefficients are significantly higher than that of ML methods (multi-step).

We reproduce \cite{guan2021explainable} that compares the performance of DRL algorithms with machine learning (ML) methods on the multi-step prediction in the portfolio allocation task. We use four technical indicators MACD, RSI, CCI, and ADX as features. Random Forest (RF), Decision Tree Regression (DT), Linear Regression (LR), and Support Vector Machine (SVM) are the ML algorithms in comparison. We use data from Dow Jones 30 constituent stocks to construct the environment. We use data from 04/01/2009 to 03/31/2020 as the training set and data from 04/01/2020 to 05/31/2022 for backtesting.

\subsection{Data Privacy, Strategy Privacy and Federated Learning Technology}

As our open-source community is continuously developing new features to ensure that FinRL-Meta provides a better user experience, one of the main targets for our next step is to enhance financial data privacy as well as strategy privacy for users. We discuss the potential of integrating the federated learning technology into our open-source FinRL-Meta, in order to achieve data privacy and strategy privacy for our users, say collaboratively training.

Federated learning is a method for training machine learning models from distributed datasets that remains private to data owners. It allows a central machine learning model to overcome the \textit{isolated data island}, i.e., to learn from data sets distributed on multiple devices that do not  reveal or share the data with a central server. We understand that in certain scenarios, our users face the problem that they only have a small amount of financial data and are not unable to train a robust model, but they are also hesitant to train their models with others due to privacy reasons. To the best of our knowledge, we would like to name a few representative examples on the use of federated learning to help financial applications. Byrd et al.~\cite{byrd2020differentially} present a privacy-preserving federated learning protocol on a real-world credit card fraud dataset for the development of federated learning systems. The researchers in WeBank~\cite{liu2021fate} created FATE, an industrial-grade project that supports enterprises and institutions to build machine learning models collaboratively at large-scale in a distributed manner. FATE has been adopted in real-world applications in finance, health and recommender systems. We also want to mention one research work~\cite{kairouz2021advances} which points out the open problems in federated learning. We believe that, as a newly introduced method, federated learning has a lot of undiscovered and exciting applications that we can develop. We would like to encourage our community members to explore the potential of federated learning technologies for the benefit of financial applications.

\subsection{FinRL-Meta and DAO, DeFi, NFT, Web3}

Over the past decades, we have witnessed capital growth exceeding economic growth globally. However, the door to personal capital growth is not open to all. In a way, one needs to start rich to get richer. The situation is even worsened by the competition with computers. Today in major stock markets, at least $60\%$ of the trades are automated by algorithms. 
    
How to \textit{democratize opportunity} for personal capital growth? 
We need to ally with the computers to take advantage of unprecedented amounts of data and unparalleled computing infrastructure. Therefore, we come up with this vision to establish an AI framework where retail traders can grow personal capital via a pay-by-use mode. Today, we have reached a full-stack solution that provides financially optimized deep reinforcement learning algorithms together with cloud-native solutions. Users can easily access numerous financial data, as well as computational resources whenever needed.

FinRL \cite{liu2020finrl,liu2021finrl} is the first open-source framework to demonstrate the great potential of financial reinforcement learning \cite{hambly2021recent}. Over several years' development\footnote{We began to build an open-source community with a practical demonstration \cite{liu2018practical} at NeurIPS 2018 conference.}, it has evolved into an ecosystem, FinRL-Meta, serving as a playground for data-driven financial reinforcement learning. We believe FinRL-Meta will reshape our financial lives, while our open-source community will make sure it is for the better.

Next, we discuss the potential of combining our open-source FinRL-Meta ecosystem with emerging technologies, such as DAO, DeFi, NFT\footnote{Ethereum: https://ethereum.org/en/}, and Web3\footnote{Web3 Foundation: https://web3.foundation/}.
\begin{itemize}
    \item Decentralized autonomous organizations (DAO) has three features: 1). Member-owned communities without centralized leadership; 2). A safe way to collaborate with Internet strangers; and 3). A safe place to commit funds to a specific cause. In FinRL-Meta's open-source community (over $12K$ users at the moment of finalizing this paper version), we plan to encourage community members to organize into DAO funds (essentially a distributed fund) in an ad hoc manner. Members of a DAO fund will employ smart contracts to crowdfunding, design strategy, implement algorithmic trading, share profit and loss, etc.
    \item Decentralized finance (DeFi) has three features: 1). A global, open alternative to the current financial system; 2). Products that let you borrow, save, invest, trade, and more; and 3). Based on open-source technology that anyone can program with. We would like to encourage community members to actively explore the potential of financial reinforcement learning technologies in the emerging DeFi based financial markets. For example, using the automated environment layer in Section \ref{sect:env_layer}, we will connect DeFi trading systems into gym-style market environments; also, we are actively working with financial data providers and cloud computing providers to  upgrade the trading infrastructure.
    \item Non-fungible token (NFT) has three features: 1). A way to represent anything unique as an Ethereum-based asset; 2). NFTs are giving more power to content creators than ever before; and 3). Powered by smart contracts on the Ethereum blockchain. In the FinRL-Meta community, we would like to encourage community members to release trading strategies and codes in the NFT forms, while a DAO fund will buy these NFTs via smart contracts and then trade in a DeFi system.
    \item Third-generation of WWW (Web3) incorporates concepts such as decentralization, blockchain technologies, and token-based economics, which is essentially community-run Internet. Our open-source FinRL-Meta community will serve Web3 users a unique infrastructure and playground, where Web3 users can establish DAO funds to trade in a DeFi system.
\end{itemize}

As a near-term development, the FinRL-Meta community would embrace the Massive Open Online Courses (MOOC) paradigm. For demonstrative and educational purposes, our community members  are actively creating notebooks, blogs and videos. We believe it would be an effective incentive to release them in the form of NFTs. Web3MOOC guarantees authorship and ownership (a right to profit) of those NFTs, thus providing a strong incentive to our community members for content creation.

\subsection{Accessibility, Accountability, Maintenance and Rights}

FinRL-Meta is an open-source project, held on GitHub. We use the MIT License for research and educational usage. while users can utilize them as stepping stones for customized trading strategies. Codes, market environments, benchmarks and documentations are available on the GitHub repository:\\ \url{https://github.com/AI4Finance-Foundation/FinRL-Meta}.

FinRL-Meta has been actively maintained by the AI4Finance community which has over $12K$ members at the moment. On GitHub, we keep updating our codes, merging pull requests, and fixing bugs and issues. We welcome contributions from community members, researchers and quant traders.

We have accumulated six competitive advantages over the past five years. The first three are technological innovations: 
\begin{itemize}
    \item FinRL \cite{liu2020finrl,liu2021finrl} is the first framework to provide an automatic pipeline for financial reinforcement learning.
    \item For financial big data, the FinRL-Meta project connects with > 30 market data sources.
    \item For cloud solutions, the FinRL-Podracer project \cite{finrl_podracer_2021} \cite{liu2021podracer} scales out to $\geq 1000$ GPUs. We have extensive testings on NVIDIA’s DGX-2 SuperPod platform.
\end{itemize}

Based on the above projects and active contributions, an open-source community in the intersection of ML and Finance fields is emerging. Our AI4Finance community is robust with the following three features:
\begin{itemize}
    \item We have over $12K$ active community members, many of which are actively designing strategies and connecting with paper trading, even live trading. We are collaborating with  tens of universities and research institutes, and $\geq 50$ software engineers from IT companies.
    \item Both Columbia University (Department of Electrical Engineering, Department of Statistics) and New York University ((Department of Finance) have opened delicate courses about FinRL, while $\geq 120$ students in total have taken it.
    \item In academia, we have several accepted papers and also delivered several invited talks. Our AI4Finance Foundation (\url{https://github.com/AI4Finance-Foundation}) serves as a bridge between machine learning, data science, operation research, and finance communities.
\end{itemize}

\section*{Acknowledgement}

We thank Mr. Tao Liu (IDEA Research, International Digital Economy Academy) for technical support of computing platform on this research project. Ming Zhu was supported by National Natural Science Foundations of China (Grant No. 61902387). Christina Dan Wang is supported in part by National Natural Science Foundation of China (NNSFC) grant 12271363 and 11901395.

\textbf{Disclaimer: Nothing in this paper and the FinRL-Meta repository is financial advice, and NOT a recommendation to trade real money. Please use common sense and always first consult a professional before trading or investing.}
\\
\\
\\



\bibliography{ref} 


\begin{thebibliography}{91}
\ifx \bisbn   \undefined \def \bisbn  #1{ISBN #1}\fi
\ifx \binits  \undefined \def \binits#1{#1}\fi
\ifx \bauthor  \undefined \def \bauthor#1{#1}\fi
\ifx \batitle  \undefined \def \batitle#1{#1}\fi
\ifx \bjtitle  \undefined \def \bjtitle#1{#1}\fi
\ifx \bvolume  \undefined \def \bvolume#1{\textbf{#1}}\fi
\ifx \byear  \undefined \def \byear#1{#1}\fi
\ifx \bissue  \undefined \def \bissue#1{#1}\fi
\ifx \bfpage  \undefined \def \bfpage#1{#1}\fi
\ifx \blpage  \undefined \def \blpage #1{#1}\fi
\ifx \burl  \undefined \def \burl#1{\textsf{#1}}\fi
\ifx \doiurl  \undefined \def \doiurl#1{\url{https://doi.org/#1}}\fi
\ifx \betal  \undefined \def \betal{\textit{et al.}}\fi
\ifx \binstitute  \undefined \def \binstitute#1{#1}\fi
\ifx \binstitutionaled  \undefined \def \binstitutionaled#1{#1}\fi
\ifx \bctitle  \undefined \def \bctitle#1{#1}\fi
\ifx \beditor  \undefined \def \beditor#1{#1}\fi
\ifx \bpublisher  \undefined \def \bpublisher#1{#1}\fi
\ifx \bbtitle  \undefined \def \bbtitle#1{#1}\fi
\ifx \bedition  \undefined \def \bedition#1{#1}\fi
\ifx \bseriesno  \undefined \def \bseriesno#1{#1}\fi
\ifx \blocation  \undefined \def \blocation#1{#1}\fi
\ifx \bsertitle  \undefined \def \bsertitle#1{#1}\fi
\ifx \bsnm \undefined \def \bsnm#1{#1}\fi
\ifx \bsuffix \undefined \def \bsuffix#1{#1}\fi
\ifx \bparticle \undefined \def \bparticle#1{#1}\fi
\ifx \barticle \undefined \def \barticle#1{#1}\fi
\bibcommenthead
\ifx \bconfdate \undefined \def \bconfdate #1{#1}\fi
\ifx \botherref \undefined \def \botherref #1{#1}\fi
\ifx \url \undefined \def \url#1{\textsf{#1}}\fi
\ifx \bchapter \undefined \def \bchapter#1{#1}\fi
\ifx \bbook \undefined \def \bbook#1{#1}\fi
\ifx \bcomment \undefined \def \bcomment#1{#1}\fi
\ifx \oauthor \undefined \def \oauthor#1{#1}\fi
\ifx \citeauthoryear \undefined \def \citeauthoryear#1{#1}\fi
\ifx \endbibitem  \undefined \def \endbibitem {}\fi
\ifx \bconflocation  \undefined \def \bconflocation#1{#1}\fi
\ifx \arxivurl  \undefined \def \arxivurl#1{\textsf{#1}}\fi
\csname PreBibitemsHook\endcsname

\bibitem{liu2021finrl}
\begin{botherref}
\oauthor{\bsnm{Liu}, \binits{X.-Y.}},
\oauthor{\bsnm{Yang}, \binits{H.}},
\oauthor{\bsnm{Gao}, \binits{J.}},
\oauthor{\bsnm{Wang}, \binits{C.D.}}:
{FinRL}: Deep reinforcement learning framework to automate trading in
  quantitative finance.
ACM International Conference on AI in Finance (ICAIF)
(2021)
\end{botherref}
\endbibitem

\bibitem{hambly2021recent}
\begin{botherref}
\oauthor{\bsnm{Hambly}, \binits{B.}},
\oauthor{\bsnm{Xu}, \binits{R.}},
\oauthor{\bsnm{Yang}, \binits{H.}}:
Recent advances in reinforcement learning in finance.
Mathematical Finance
(2023)
\end{botherref}
\endbibitem

\bibitem{sutton2018reinforcement}
\begin{bbook}
\bauthor{\bsnm{Sutton}, \binits{R.S.}},
\bauthor{\bsnm{Barto}, \binits{A.G.}}:
\bbtitle{Reinforcement Learning: An Introduction},
(\byear{2018})
\end{bbook}
\endbibitem

\bibitem{DQN}
\begin{barticle}
\bauthor{\bsnm{Mnih}, \binits{V.}},
\bauthor{\bsnm{Kavukcuoglu}, \binits{K.}},
\bauthor{\bsnm{Silver}, \binits{D.}},
\bauthor{\bsnm{Rusu}, \binits{A.}},
\bauthor{\bsnm{Veness}, \binits{J.}},
\bauthor{\bsnm{Bellemare}, \binits{M.}},
\bauthor{\bsnm{Graves}, \binits{A.}},
\bauthor{\bsnm{Riedmiller}, \binits{M.}},
\bauthor{\bsnm{Fidjeland}, \binits{A.}},
\bauthor{\bsnm{Ostrovski}, \binits{G.}},
\bauthor{\bsnm{Petersen}, \binits{S.}},
\bauthor{\bsnm{Beattie}, \binits{C.}},
\bauthor{\bsnm{Sadik}, \binits{A.}},
\bauthor{\bsnm{Antonoglou}, \binits{I.}},
\bauthor{\bsnm{King}, \binits{H.}},
\bauthor{\bsnm{Kumaran}, \binits{D.}},
\bauthor{\bsnm{Wierstra}, \binits{D.}},
\bauthor{\bsnm{Legg}, \binits{S.}},
\bauthor{\bsnm{Hassabis}, \binits{D.}}:
\batitle{Human-level control through deep reinforcement learning}.
\bjtitle{Nature}
\bvolume{518},
\bfpage{529}--\blpage{33}
(\byear{2015})
\end{barticle}
\endbibitem

\bibitem{silver2016alphaGo1}
\begin{barticle}
\bauthor{\bsnm{Silver}, \binits{D.}},
\bauthor{\bsnm{Huang}, \binits{A.}},
\bauthor{\bsnm{Maddison}, \binits{C.J.}},
\bauthor{\bsnm{Guez}, \binits{A.}},
\bauthor{\bsnm{Sifre}, \binits{L.}},
\bauthor{\bsnm{Van Den~Driessche}, \binits{G.}},
\bauthor{\bsnm{Schrittwieser}, \binits{J.}},
\bauthor{\bsnm{Antonoglou}, \binits{I.}},
\bauthor{\bsnm{Panneershelvam}, \binits{V.}},
\bauthor{\bsnm{Lanctot}, \binits{M.}}, \betal:
\batitle{Mastering the game of go with deep neural networks and tree search}.
\bjtitle{nature}
\bvolume{529}(\bissue{7587}),
\bfpage{484}--\blpage{489}
(\byear{2016})
\end{barticle}
\endbibitem

\bibitem{silver2017alphaGo2}
\begin{barticle}
\bauthor{\bsnm{Silver}, \binits{D.}},
\bauthor{\bsnm{Schrittwieser}, \binits{J.}},
\bauthor{\bsnm{Simonyan}, \binits{K.}},
\bauthor{\bsnm{Antonoglou}, \binits{I.}},
\bauthor{\bsnm{Huang}, \binits{A.}},
\bauthor{\bsnm{Guez}, \binits{A.}},
\bauthor{\bsnm{Hubert}, \binits{T.}},
\bauthor{\bsnm{Baker}, \binits{L.}},
\bauthor{\bsnm{Lai}, \binits{M.}},
\bauthor{\bsnm{Bolton}, \binits{A.}}, \betal:
\batitle{Mastering the game of {Go} without human knowledge}.
\bjtitle{Nature}
\bvolume{550}(\bissue{7676}),
\bfpage{354}--\blpage{359}
(\byear{2017})
\end{barticle}
\endbibitem

\bibitem{vinyals2019grandmaster}
\begin{barticle}
\bauthor{\bsnm{Vinyals}, \binits{O.}},
\bauthor{\bsnm{Babuschkin}, \binits{I.}},
\bauthor{\bsnm{Czarnecki}, \binits{W.M.}},
\bauthor{\bsnm{Mathieu}, \binits{M.}},
\bauthor{\bsnm{Dudzik}, \binits{A.}},
\bauthor{\bsnm{Chung}, \binits{J.}},
\bauthor{\bsnm{Choi}, \binits{D.H.}},
\bauthor{\bsnm{Powell}, \binits{R.}},
\bauthor{\bsnm{Ewalds}, \binits{T.}},
\bauthor{\bsnm{Georgiev}, \binits{P.}}, \betal:
\batitle{Grandmaster level in {StarCraft II} using multi-agent reinforcement
  learning}.
\bjtitle{Nature}
\bvolume{575}(\bissue{7782}),
\bfpage{350}--\blpage{354}
(\byear{2019})
\end{barticle}
\endbibitem

\bibitem{ouyang2022training}
\begin{barticle}
\bauthor{\bsnm{Ouyang}, \binits{L.}},
\bauthor{\bsnm{Wu}, \binits{J.}},
\bauthor{\bsnm{Jiang}, \binits{X.}},
\bauthor{\bsnm{Almeida}, \binits{D.}},
\bauthor{\bsnm{Wainwright}, \binits{C.}},
\bauthor{\bsnm{Mishkin}, \binits{P.}},
\bauthor{\bsnm{Zhang}, \binits{C.}},
\bauthor{\bsnm{Agarwal}, \binits{S.}},
\bauthor{\bsnm{Slama}, \binits{K.}},
\bauthor{\bsnm{Ray}, \binits{A.}}, \betal:
\batitle{Training language models to follow instructions with human feedback}.
\bjtitle{Advances in Neural Information Processing Systems}
\bvolume{35},
\bfpage{27730}--\blpage{27744}
(\byear{2022})
\end{barticle}
\endbibitem

\bibitem{GPT4}
\begin{botherref}
\oauthor{\bsnm{OpenAI}}:
{GPT-4} technical report.
https://arxiv.org/abs/2303.08774
(2023)
\end{botherref}
\endbibitem

\bibitem{deng2009imagenet}
\begin{bchapter}
\bauthor{\bsnm{Deng}, \binits{J.}},
\bauthor{\bsnm{Dong}, \binits{W.}},
\bauthor{\bsnm{Socher}, \binits{R.}},
\bauthor{\bsnm{Li}, \binits{L.-J.}},
\bauthor{\bsnm{Li}, \binits{K.}},
\bauthor{\bsnm{Fei-Fei}, \binits{L.}}:
\bctitle{{ImageNet}: A large-scale hierarchical image database}.
In: \bbtitle{IEEE Conference on Computer Vision and Pattern Recognition},
pp. \bfpage{248}--\blpage{255}
(\byear{2009})
\end{bchapter}
\endbibitem

\bibitem{lussange2021modelling}
\begin{barticle}
\bauthor{\bsnm{Lussange}, \binits{J.}},
\bauthor{\bsnm{Lazarevich}, \binits{I.}},
\bauthor{\bsnm{Bourgeois-Gironde}, \binits{S.}},
\bauthor{\bsnm{Palminteri}, \binits{S.}},
\bauthor{\bsnm{Gutkin}, \binits{B.}}:
\batitle{Modelling stock markets by multi-agent reinforcement learning}.
\bjtitle{Computational Economics}
\bvolume{57}(\bissue{1}),
\bfpage{113}--\blpage{147}
(\byear{2021})
\end{barticle}
\endbibitem

\bibitem{pricope2021deep}
\begin{botherref}
\oauthor{\bsnm{Pricope}, \binits{T.-V.}}:
Deep reinforcement learning in quantitative algorithmic trading: A review.
arXiv preprint arXiv:2106.00123
(2021)
\end{botherref}
\endbibitem

\bibitem{raberto2001agent}
\begin{barticle}
\bauthor{\bsnm{Raberto}, \binits{M.}},
\bauthor{\bsnm{Cincotti}, \binits{S.}},
\bauthor{\bsnm{Focardi}, \binits{S.M.}},
\bauthor{\bsnm{Marchesi}, \binits{M.}}:
\batitle{Agent-based simulation of a financial market}.
\bjtitle{Physica A: Statistical Mechanics and its Applications}
\bvolume{299}(\bissue{1-2}),
\bfpage{319}--\blpage{327}
(\byear{2001})
\end{barticle}
\endbibitem

\bibitem{liu2018practical}
\begin{botherref}
\oauthor{\bsnm{Liu}, \binits{X.-Y.}},
\oauthor{\bsnm{Xiong}, \binits{Z.}},
\oauthor{\bsnm{Zhong}, \binits{S.}},
\oauthor{\bsnm{Yang}, \binits{H.}},
\oauthor{\bsnm{Walid}, \binits{A.}}:
Practical deep reinforcement learning approach for stock trading.
Workshop on Challenges and Opportunities for AI in Financial Services, NeurIPS
(2018)
\end{botherref}
\endbibitem

\bibitem{yang2020deep}
\begin{botherref}
\oauthor{\bsnm{Yang}, \binits{H.}},
\oauthor{\bsnm{Liu}, \binits{X.-Y.}},
\oauthor{\bsnm{Zhong}, \binits{S.}},
\oauthor{\bsnm{Walid}, \binits{A.}}:
Deep reinforcement learning for automated stock trading: An ensemble strategy.
ACM International Conference on AI in Finance
(2020)
\end{botherref}
\endbibitem

\bibitem{zhang2020deep}
\begin{barticle}
\bauthor{\bsnm{Zhang}, \binits{Z.}},
\bauthor{\bsnm{Zohren}, \binits{S.}},
\bauthor{\bsnm{Roberts}, \binits{S.}}:
\batitle{Deep reinforcement learning for trading}.
\bjtitle{The Journal of Financial Data Science}
\bvolume{2}(\bissue{2}),
\bfpage{25}--\blpage{40}
(\byear{2020})
\end{barticle}
\endbibitem

\bibitem{bao2019multiagent}
\begin{botherref}
\oauthor{\bsnm{Bao}, \binits{W.}},
\oauthor{\bsnm{Liu}, \binits{X.-Y.}}:
Multi-agent deep reinforcement learning for liquidation strategy analysis.
ICML Workshop on Applications and Infrastructure for Multi-Agent Learning
(2019)
\end{botherref}
\endbibitem

\bibitem{amrouni2021abides}
\begin{botherref}
\oauthor{\bsnm{Amrouni}, \binits{S.}},
\oauthor{\bsnm{Moulin}, \binits{A.}},
\oauthor{\bsnm{Vann}, \binits{J.}},
\oauthor{\bsnm{Vyetrenko}, \binits{S.}},
\oauthor{\bsnm{Balch}, \binits{T.}},
\oauthor{\bsnm{Veloso}, \binits{M.}}:
{ABIDES-Gym}: Gym environments for multi-agent discrete event simulation and
  application to financial markets.
ACM International Conference on AI in Finance (ICAIF)
(2021)
\end{botherref}
\endbibitem

\bibitem{market_simulator}
\begin{botherref}
\oauthor{\bsnm{Han}, \binits{J.}},
\oauthor{\bsnm{Xia}, \binits{Z.}},
\oauthor{\bsnm{Liu}, \binits{X.-Y.}},
\oauthor{\bsnm{Zhang}, \binits{C.}},
\oauthor{\bsnm{Wang}, \binits{Z.}},
\oauthor{\bsnm{Guo}, \binits{j.}}:
Massively parallel market simulator for financial reinforcement learning.
AI in Finance Bridge, AAAI
(2023)
\end{botherref}
\endbibitem

\bibitem{liu2020finrl}
\begin{botherref}
\oauthor{\bsnm{Liu}, \binits{X.-Y.}},
\oauthor{\bsnm{Yang}, \binits{H.}},
\oauthor{\bsnm{Chen}, \binits{Q.}},
\oauthor{\bsnm{Zhang}, \binits{R.}},
\oauthor{\bsnm{Yang}, \binits{L.}},
\oauthor{\bsnm{Xiao}, \binits{B.}},
\oauthor{\bsnm{Wang}, \binits{C.D.}}:
{FinRL}: A deep reinforcement learning library for automated stock trading in
  quantitative finance.
Deep RL Workshop, NeurIPS
(2020)
\end{botherref}
\endbibitem

\bibitem{liu2022finrlmeta}
\begin{botherref}
\oauthor{\bsnm{Liu}, \binits{X.-Y.}},
\oauthor{\bsnm{Xia}, \binits{Z.}},
\oauthor{\bsnm{Rui}, \binits{J.}},
\oauthor{\bsnm{Gao}, \binits{J.}},
\oauthor{\bsnm{Yang}, \binits{H.}},
\oauthor{\bsnm{Zhu}, \binits{M.}},
\oauthor{\bsnm{Wang}, \binits{C.D.}},
\oauthor{\bsnm{Wang}, \binits{Z.}},
\oauthor{\bsnm{Guo}, \binits{J.}}:
{FinRL-Meta}: Market environments and benchmarks for data-driven financial
  reinforcement learning.
In: Thirty-sixth Conference on Neural Information Processing Systems
\end{botherref}
\endbibitem

\bibitem{DulacArnold2020AnEI}
\begin{barticle}
\bauthor{\bsnm{Dulac-Arnold}, \binits{G.}},
\bauthor{\bsnm{Levine}, \binits{N.}},
\bauthor{\bsnm{Mankowitz}, \binits{D.J.}},
\bauthor{\bsnm{Li}, \binits{J.}},
\bauthor{\bsnm{Paduraru}, \binits{C.}},
\bauthor{\bsnm{Gowal}, \binits{S.}},
\bauthor{\bsnm{Hester}, \binits{T.}}:
\batitle{Challenges of real-world reinforcement learning: definitions,
  benchmarks and analysis}.
\bjtitle{Machine Learning}
\bvolume{110}(\bissue{9}),
\bfpage{2419}--\blpage{2468}
(\byear{2021})
\end{barticle}
\endbibitem

\bibitem{dulac2019challenges}
\begin{botherref}
\oauthor{\bsnm{Dulac-Arnold}, \binits{G.}},
\oauthor{\bsnm{Mankowitz}, \binits{D.}},
\oauthor{\bsnm{Hester}, \binits{T.}}:
Challenges of real-world reinforcement learning.
ICML Workshop on Reinforcement Learning for Real Life
(2019)
\end{botherref}
\endbibitem

\bibitem{whang2023data}
\begin{botherref}
\oauthor{\bsnm{Whang}, \binits{S.E.}},
\oauthor{\bsnm{Roh}, \binits{Y.}},
\oauthor{\bsnm{Song}, \binits{H.}},
\oauthor{\bsnm{Lee}, \binits{J.-G.}}:
Data collection and quality challenges in deep learning: A data-centric {AI}
  perspective.
The VLDB Journal,
1--23
(2023)
\end{botherref}
\endbibitem

\bibitem{zha2023data}
\begin{botherref}
\oauthor{\bsnm{Zha}, \binits{D.}},
\oauthor{\bsnm{Bhat}, \binits{Z.P.}},
\oauthor{\bsnm{Lai}, \binits{K.-H.}},
\oauthor{\bsnm{Yang}, \binits{F.}},
\oauthor{\bsnm{Jiang}, \binits{Z.}},
\oauthor{\bsnm{Zhong}, \binits{S.}},
\oauthor{\bsnm{Hu}, \binits{X.}}:
Data-centric artificial intelligence: A survey.
arXiv preprint arXiv:2303.10158
(2023)
\end{botherref}
\endbibitem

\bibitem{zha2023data2}
\begin{botherref}
\oauthor{\bsnm{Zha}, \binits{D.}},
\oauthor{\bsnm{Bhat}, \binits{Z.P.}},
\oauthor{\bsnm{Lai}, \binits{K.-H.}},
\oauthor{\bsnm{Yang}, \binits{F.}},
\oauthor{\bsnm{Hu}, \binits{X.}}:
Data-centric ai: Perspectives and challenges.
arXiv preprint arXiv:2301.04819
(2023)
\end{botherref}
\endbibitem

\bibitem{atwal2019practical}
\begin{bbook}
\bauthor{\bsnm{Atwal}, \binits{H.}}:
\bbtitle{Practical {DataOps}: Delivering Agile Data Science at Scale},
(\byear{2019})
\end{bbook}
\endbibitem

\bibitem{ereth2018dataops}
\begin{barticle}
\bauthor{\bsnm{Ereth}, \binits{J.}}:
\batitle{{DataOps}: Towards a definition.}
\bjtitle{LWDA}
\bvolume{2191},
\bfpage{104}--\blpage{112}
(\byear{2018})
\end{barticle}
\endbibitem

\bibitem{mazumder2022dataperf}
\begin{botherref}
\oauthor{\bsnm{Mazumder}, \binits{M.}},
\oauthor{\bsnm{Banbury}, \binits{C.}},
\oauthor{\bsnm{Yao}, \binits{X.}},
\oauthor{\bsnm{Karla{\v{s}}}, \binits{B.}},
\oauthor{\bsnm{Rojas}, \binits{W.G.}},
\oauthor{\bsnm{Diamos}, \binits{S.}},
\oauthor{\bsnm{Diamos}, \binits{G.}},
\oauthor{\bsnm{He}, \binits{L.}},
\oauthor{\bsnm{Kiela}, \binits{D.}},
\oauthor{\bsnm{Jurado}, \binits{D.}}, et al.:
Dataperf: Benchmarks for data-centric ai development.
arXiv preprint arXiv:2207.10062
(2022)
\end{botherref}
\endbibitem

\bibitem{sambasivan2021everyone}
\begin{bchapter}
\bauthor{\bsnm{Sambasivan}, \binits{N.}},
\bauthor{\bsnm{Kapania}, \binits{S.}},
\bauthor{\bsnm{Highfill}, \binits{H.}},
\bauthor{\bsnm{Akrong}, \binits{D.}},
\bauthor{\bsnm{Paritosh}, \binits{P.}},
\bauthor{\bsnm{Aroyo}, \binits{L.M.}}:
\bctitle{“everyone wants to do the model work, not the data work”: Data
  cascades in high-stakes ai}.
In: \bbtitle{Proceedings of the 2021 CHI Conference on Human Factors in
  Computing Systems},
pp. \bfpage{1}--\blpage{15}
(\byear{2021})
\end{bchapter}
\endbibitem

\bibitem{polyzotis2021can}
\begin{botherref}
\oauthor{\bsnm{Polyzotis}, \binits{N.}},
\oauthor{\bsnm{Zaharia}, \binits{M.}}:
What can data-centric ai learn from data and ml engineering?
arXiv preprint arXiv:2112.06439
(2021)
\end{botherref}
\endbibitem

\bibitem{ardon2021towards}
\begin{botherref}
\oauthor{\bsnm{Ardon}, \binits{L.}},
\oauthor{\bsnm{Vadori}, \binits{N.}},
\oauthor{\bsnm{Spooner}, \binits{T.}},
\oauthor{\bsnm{Xu}, \binits{M.}},
\oauthor{\bsnm{Vann}, \binits{J.}},
\oauthor{\bsnm{Ganesh}, \binits{S.}}:
Towards a fully {RL}-based market simulator.
ACM International Conference on AI in Finance (ICAIF)
(2021)
\end{botherref}
\endbibitem

\bibitem{coletta2021towards}
\begin{botherref}
\oauthor{\bsnm{Coletta}, \binits{A.}},
\oauthor{\bsnm{Prata}, \binits{M.}},
\oauthor{\bsnm{Conti}, \binits{M.}},
\oauthor{\bsnm{Mercanti}, \binits{E.}},
\oauthor{\bsnm{Bartolini}, \binits{N.}},
\oauthor{\bsnm{Moulin}, \binits{A.}},
\oauthor{\bsnm{Vyetrenko}, \binits{S.}},
\oauthor{\bsnm{Balch}, \binits{T.}}:
Towards realistic market simulations: a generative adversarial networks
  approach.
ACM International Conference on AI in Finance (ICAIF)
(2021)
\end{botherref}
\endbibitem

\bibitem{levine2020offline}
\begin{botherref}
\oauthor{\bsnm{Levine}, \binits{S.}},
\oauthor{\bsnm{Kumar}, \binits{A.}},
\oauthor{\bsnm{Tucker}, \binits{G.}},
\oauthor{\bsnm{Fu}, \binits{J.}}:
Offline reinforcement learning: Tutorial, review, and perspectives on open
  problems.
arXiv preprint arXiv:2005.01643
(2020)
\end{botherref}
\endbibitem

\bibitem{jpmorgan2023asset}
\begin{botherref}
\oauthor{\bsnm{Mahfouz}, \binits{M.}},
\oauthor{\bsnm{Gopalakrishnan}, \binits{S.}},
\oauthor{\bsnm{Suau}, \binits{M.}},
\oauthor{\bsnm{Patra}, \binits{S.}},
\oauthor{\bsnm{P.~Mandic}, \binits{D.}},
\oauthor{\bsnm{Magazzeni}, \binits{D.}},
\oauthor{\bsnm{Veloso}, \binits{M.}}:
Towards asset allocation using behavioural cloning and reinforcement learning.
AAAI AI for Financial Services Bridge
(2023)
\end{botherref}
\endbibitem

\bibitem{christiano2017deep}
\begin{botherref}
\oauthor{\bsnm{Christiano}, \binits{P.F.}},
\oauthor{\bsnm{Leike}, \binits{J.}},
\oauthor{\bsnm{Brown}, \binits{T.}},
\oauthor{\bsnm{Martic}, \binits{M.}},
\oauthor{\bsnm{Legg}, \binits{S.}},
\oauthor{\bsnm{Amodei}, \binits{D.}}:
Deep reinforcement learning from human preferences.
Advances in Neural Information Processing Systems
\textbf{30}
(2017)
\end{botherref}
\endbibitem

\bibitem{brockman2016openai}
\begin{botherref}
\oauthor{\bsnm{Brockman}, \binits{G.}},
\oauthor{\bsnm{Cheung}, \binits{V.}},
\oauthor{\bsnm{Pettersson}, \binits{L.}},
\oauthor{\bsnm{Schneider}, \binits{J.}},
\oauthor{\bsnm{Schulman}, \binits{J.}},
\oauthor{\bsnm{Tang}, \binits{J.}},
\oauthor{\bsnm{Zaremba}, \binits{W.}}:
{OpenAI Gym}.
arXiv preprint arXiv:1606.01540
(2016)
\end{botherref}
\endbibitem

\bibitem{stable-baselines}
\begin{botherref}
\oauthor{\bsnm{Raffin}, \binits{A.}},
\oauthor{\bsnm{Hill}, \binits{A.}},
\oauthor{\bsnm{Gleave}, \binits{A.}},
\oauthor{\bsnm{Kanervisto}, \binits{A.}},
\oauthor{\bsnm{Ernestus}, \binits{M.}},
\oauthor{\bsnm{Dormann}, \binits{N.}}:
Stable-baselines3: Reliable reinforcement learning implementations.
Journal of Machine Learning Research
(2021)
\end{botherref}
\endbibitem

\bibitem{liang2018rllib}
\begin{bchapter}
\bauthor{\bsnm{Liang}, \binits{E.}},
\bauthor{\bsnm{Liaw}, \binits{R.}},
\bauthor{\bsnm{Nishihara}, \binits{R.}},
\bauthor{\bsnm{Moritz}, \binits{P.}},
\bauthor{\bsnm{Fox}, \binits{R.}},
\bauthor{\bsnm{Goldberg}, \binits{K.}},
\bauthor{\bsnm{Gonzalez}, \binits{J.}},
\bauthor{\bsnm{Jordan}, \binits{M.}},
\bauthor{\bsnm{Stoica}, \binits{I.}}:
\bctitle{{RLlib}: Abstractions for distributed reinforcement learning}.
In: \bbtitle{International Conference on Machine Learning},
pp. \bfpage{3053}--\blpage{3062}
(\byear{2018}).
\bcomment{PMLR}
\end{bchapter}
\endbibitem

\bibitem{elegantrl}
\begin{botherref}
\oauthor{\bsnm{Liu}, \binits{X.-Y.}},
\oauthor{\bsnm{Li}, \binits{Z.}},
\oauthor{\bsnm{Wang}, \binits{Z.}},
\oauthor{\bsnm{Zheng}, \binits{J.}}:
{ElegantRL}: A Lightweight and Stable Deep Reinforcement Learning Library.
GitHub
(2021)
\end{botherref}
\endbibitem

\bibitem{fu2020d4rl}
\begin{botherref}
\oauthor{\bsnm{Fu}, \binits{J.}},
\oauthor{\bsnm{Kumar}, \binits{A.}},
\oauthor{\bsnm{Nachum}, \binits{O.}},
\oauthor{\bsnm{Tucker}, \binits{G.}},
\oauthor{\bsnm{Levine}, \binits{S.}}:
{D4RL}: Datasets for deep data-driven reinforcement learning.
arXiv preprint arXiv:2004.07219
(2020)
\end{botherref}
\endbibitem

\bibitem{qin2021neorl}
\begin{botherref}
\oauthor{\bsnm{Qin}, \binits{R.}},
\oauthor{\bsnm{Gao}, \binits{S.}},
\oauthor{\bsnm{Zhang}, \binits{X.}},
\oauthor{\bsnm{Xu}, \binits{Z.}},
\oauthor{\bsnm{Huang}, \binits{S.}},
\oauthor{\bsnm{Li}, \binits{Z.}},
\oauthor{\bsnm{Zhang}, \binits{W.}},
\oauthor{\bsnm{Yu}, \binits{Y.}}:
{NeoRL}: A near real-world benchmark for offline reinforcement learning.
NeurIPS {D}atasets and {B}enchmarks
(2022)
\end{botherref}
\endbibitem

\bibitem{vazquez2019cityLearn}
\begin{botherref}
\oauthor{\bsnm{V{\'a}zquez-Canteli}, \binits{J.R.}},
\oauthor{\bsnm{K{\"a}mpf}, \binits{J.}},
\oauthor{\bsnm{Henze}, \binits{G.}},
\oauthor{\bsnm{Nagy}, \binits{Z.}}:
{CityLearn v1.0}: An {OpenAI} gym environment for demand response with deep
  reinforcement learning.
ACM International Conference on Systems for Energy-Efficient Buildings, Cities,
  and Transportation
(2019)
\end{botherref}
\endbibitem

\bibitem{hein2017benchmark}
\begin{bchapter}
\bauthor{\bsnm{Hein}, \binits{D.}},
\bauthor{\bsnm{Depeweg}, \binits{S.}},
\bauthor{\bsnm{Tokic}, \binits{M.}},
\bauthor{\bsnm{Udluft}, \binits{S.}},
\bauthor{\bsnm{Hentschel}, \binits{A.}},
\bauthor{\bsnm{Runkler}, \binits{T.A.}},
\bauthor{\bsnm{Sterzing}, \binits{V.}}:
\bctitle{A benchmark environment motivated by industrial control problems}.
In: \bbtitle{IEEE Symposium Series on Computational Intelligence (SSCI)},
pp. \bfpage{1}--\blpage{8}
(\byear{2017}).
\bcomment{IEEE}
\end{bchapter}
\endbibitem

\bibitem{todorov2012mujoco}
\begin{bchapter}
\bauthor{\bsnm{Todorov}, \binits{E.}},
\bauthor{\bsnm{Erez}, \binits{T.}},
\bauthor{\bsnm{Tassa}, \binits{Y.}}:
\bctitle{{MuJoCo}: A physics engine for model-based control}.
In: \bbtitle{IEEE/RSJ {International Conference on Intelligent Robots and
  Systems}},
pp. \bfpage{5026}--\blpage{5033}
(\byear{2012}).
\bcomment{IEEE}
\end{bchapter}
\endbibitem

\bibitem{Treleaven2013AlgorithmicTR}
\begin{barticle}
\bauthor{\bsnm{Treleaven}, \binits{P.}},
\bauthor{\bsnm{Galas}, \binits{M.}},
\bauthor{\bsnm{Lalchand}, \binits{V.}}:
\batitle{Algorithmic trading review}.
\bjtitle{Communications of the ACM}
\bvolume{56},
\bfpage{76}--\blpage{85}
(\byear{2013})
\end{barticle}
\endbibitem

\bibitem{Nuti2011AlgorithmicT}
\begin{barticle}
\bauthor{\bsnm{Nuti}, \binits{G.}},
\bauthor{\bsnm{Mirghaemi}, \binits{M.}},
\bauthor{\bsnm{Treleaven}, \binits{P.}},
\bauthor{\bsnm{Yingsaeree}, \binits{C.}}:
\batitle{Algorithmic trading}.
\bjtitle{Computer}
\bvolume{44},
\bfpage{61}--\blpage{69}
(\byear{2011})
\end{barticle}
\endbibitem

\bibitem{alla2021mlops}
\begin{bchapter}
\bauthor{\bsnm{Alla}, \binits{S.}},
\bauthor{\bsnm{Adari}, \binits{S.K.}}:
\bctitle{What is {MLOps?}}
In: \bbtitle{Beginning MLOps with MLFlow},
pp. \bfpage{79}--\blpage{124}
(\byear{2021})
\end{bchapter}
\endbibitem

\bibitem{wilkman2020feasibility}
\begin{botherref}
\oauthor{\bsnm{Wilkman}, \binits{M.}}, et al.:
Feasibility of a reinforcement learning based stock trader.
Aaltodoc
(2020)
\end{botherref}
\endbibitem

\bibitem{brown1992survivorship}
\begin{barticle}
\bauthor{\bsnm{Brown}, \binits{S.J.}},
\bauthor{\bsnm{Goetzmann}, \binits{W.}},
\bauthor{\bsnm{Ibbotson}, \binits{R.G.}},
\bauthor{\bsnm{Ross}, \binits{S.A.}}:
\batitle{Survivorship bias in performance studies}.
\bjtitle{The Review of Financial Studies}
\bvolume{5}(\bissue{4}),
\bfpage{553}--\blpage{580}
(\byear{1992})
\end{barticle}
\endbibitem

\bibitem{gort2022deep}
\begin{botherref}
\oauthor{\bsnm{Gort}, \binits{B.}},
\oauthor{\bsnm{Liu}, \binits{X.-Y.}},
\oauthor{\bsnm{Sun}, \binits{X.}},
\oauthor{\bsnm{Gao}, \binits{J.}},
\oauthor{\bsnm{Chen}, \binits{S.}},
\oauthor{\bsnm{Wang}, \binits{C.D.}}:
Deep reinforcement learning for cryptocurrency trading: Practical approach to
  address backtest overfitting.
AI in Finance Bridge, AAAI
(2023)
\end{botherref}
\endbibitem

\bibitem{de2018advances}
\begin{bbook}
\bauthor{\bsnm{De~Prado}, \binits{M.L.}}:
\bbtitle{Advances in Financial Machine Learning},
(\byear{2018})
\end{bbook}
\endbibitem

\bibitem{mamon2007hidden}
\begin{bbook}
\bauthor{\bsnm{Mamon}, \binits{R.S.}},
\bauthor{\bsnm{Elliott}, \binits{R.J.}}:
\bbtitle{Hidden Markov Models in Finance}
vol. \bseriesno{4},
(\byear{2007})
\end{bbook}
\endbibitem

\bibitem{liu2020adaptive}
\begin{bchapter}
\bauthor{\bsnm{Liu}, \binits{Y.}},
\bauthor{\bsnm{Liu}, \binits{Q.}},
\bauthor{\bsnm{Zhao}, \binits{H.}},
\bauthor{\bsnm{Pan}, \binits{Z.}},
\bauthor{\bsnm{Liu}, \binits{C.}}:
\bctitle{Adaptive quantitative trading: An imitative deep reinforcement
  learning approach}.
In: \bbtitle{Proceedings of the AAAI {C}onference on {A}rtificial
  {I}ntelligence)},
vol. \bseriesno{34},
pp. \bfpage{2128}--\blpage{2135}
(\byear{2020})
\end{bchapter}
\endbibitem

\bibitem{rundo2019deep}
\begin{barticle}
\bauthor{\bsnm{Rundo}, \binits{F.}}:
\batitle{Deep {LSTM} with reinforcement learning layer for financial trend
  prediction in fx high frequency trading systems}.
\bjtitle{Applied Sciences}
\bvolume{9}(\bissue{20}),
\bfpage{4460}
(\byear{2019})
\end{barticle}
\endbibitem

\bibitem{xiao2020feature}
\begin{barticle}
\bauthor{\bsnm{Xiao}, \binits{G.}},
\bauthor{\bsnm{Li}, \binits{J.}},
\bauthor{\bsnm{Chen}, \binits{Y.}},
\bauthor{\bsnm{Li}, \binits{K.}}:
\batitle{Malfcs: An effective malware classification framework with automated
  feature extraction based on deep convolutional neural networks}.
\bjtitle{Elsevier Journal of Parallel and Distributed Computing}
\bvolume{141},
\bfpage{49}--\blpage{58}
(\byear{2020})
\end{barticle}
\endbibitem

\bibitem{nargesian2017Feature}
\begin{bchapter}
\bauthor{\bsnm{Nargesian}, \binits{F.}},
\bauthor{\bsnm{Samulowitz}, \binits{H.}},
\bauthor{\bsnm{Khurana}, \binits{U.}},
\bauthor{\bsnm{Khalil}, \binits{E.B.}},
\bauthor{\bsnm{Turaga}, \binits{D.S.}}:
\bctitle{Learning feature engineering for classification.}
In: \bbtitle{IJCAI},
vol. \bseriesno{17},
pp. \bfpage{2529}--\blpage{2535}
(\byear{2017})
\end{bchapter}
\endbibitem

\bibitem{xinyi_2019}
\begin{botherref}
\oauthor{\bsnm{Li}, \binits{X.}},
\oauthor{\bsnm{Li}, \binits{Y.}},
\oauthor{\bsnm{Yang}, \binits{H.}},
\oauthor{\bsnm{Yang}, \binits{L.}},
\oauthor{\bsnm{Liu}, \binits{X.-Y.}}:
{DP-LSTM}: Differential privacy-inspired lstm for stock prediction using
  financial news.
33rd Conference on Neural Information Processing Systems Workshop on Robust AI
  in Financial Services: Data, Fairness, Explainability, Trustworthiness, and
  Privacy, December 2019
(2019)
\end{botherref}
\endbibitem

\bibitem{fang2019practical}
\begin{bchapter}
\bauthor{\bsnm{Fang}, \binits{Y.}},
\bauthor{\bsnm{Liu}, \binits{X.-Y.}},
\bauthor{\bsnm{Yang}, \binits{H.}}:
\bctitle{Practical machine learning approach to capture the scholar data driven
  {Alpha} in {AI} industry}.
In: \bbtitle{IEEE International Conference on Big Data (Big Data)},
pp. \bfpage{2230}--\blpage{2239}
(\byear{2019}).
\bcomment{IEEE}
\end{bchapter}
\endbibitem

\bibitem{chen2020quantifying}
\begin{bchapter}
\bauthor{\bsnm{Chen}, \binits{Q.}},
\bauthor{\bsnm{Liu}, \binits{X.-Y.}}:
\bctitle{Quantifying {ESG} alpha using scholar big data: an automated machine
  learning approach}.
In: \bbtitle{Proceedings of the First ACM International Conference on AI in
  Finance},
pp. \bfpage{1}--\blpage{8}
(\byear{2020})
\end{bchapter}
\endbibitem

\bibitem{xing2018natural}
\begin{barticle}
\bauthor{\bsnm{Xing}, \binits{F.Z.}},
\bauthor{\bsnm{Cambria}, \binits{E.}},
\bauthor{\bsnm{Welsch}, \binits{R.E.}}:
\batitle{Natural language based financial forecasting: a survey}.
\bjtitle{Artificial Intelligence Review}
\bvolume{50}(\bissue{1}),
\bfpage{49}--\blpage{73}
(\byear{2018})
\end{barticle}
\endbibitem

\bibitem{loughran2011liability}
\begin{barticle}
\bauthor{\bsnm{Loughran}, \binits{T.}},
\bauthor{\bsnm{McDonald}, \binits{B.}}:
\batitle{When is a liability not a liability? textual analysis, dictionaries,
  and 10-ks}.
\bjtitle{The Journal of finance}
\bvolume{66}(\bissue{1}),
\bfpage{35}--\blpage{65}
(\byear{2011})
\end{barticle}
\endbibitem

\bibitem{hamilton2016inducing}
\begin{bchapter}
\bauthor{\bsnm{Hamilton}, \binits{W.L.}},
\bauthor{\bsnm{Clark}, \binits{K.}},
\bauthor{\bsnm{Leskovec}, \binits{J.}},
\bauthor{\bsnm{Jurafsky}, \binits{D.}}:
\bctitle{Inducing domain-specific sentiment lexicons from unlabeled corpora}.
In: \bbtitle{Proceedings of the Conference on Empirical Methods in Natural
  Language Processing. Conference on Empirical Methods in Natural Language
  Processing},
vol. \bseriesno{2016},
p. \bfpage{595}
(\byear{2016}).
\bcomment{NIH Public Access}
\end{bchapter}
\endbibitem

\bibitem{tai2013automatic}
\begin{bchapter}
\bauthor{\bsnm{Tai}, \binits{Y.-J.}},
\bauthor{\bsnm{Kao}, \binits{H.-Y.}}:
\bctitle{Automatic domain-specific sentiment lexicon generation with label
  propagation}.
In: \bbtitle{Proceedings of International Conference on Information Integration
  and Web-based Applications \& Services},
pp. \bfpage{53}--\blpage{62}
(\byear{2013})
\end{bchapter}
\endbibitem

\bibitem{hutto2014vader}
\begin{bchapter}
\bauthor{\bsnm{Hutto}, \binits{C.}},
\bauthor{\bsnm{Gilbert}, \binits{E.}}:
\bctitle{Vader: A parsimonious rule-based model for sentiment analysis of
  social media text}.
In: \bbtitle{Proceedings of the International AAAI Conference on Web and Social
  Media},
vol. \bseriesno{8},
pp. \bfpage{216}--\blpage{225}
(\byear{2014})
\end{bchapter}
\endbibitem

\bibitem{chen2018ntusd}
\begin{bchapter}
\bauthor{\bsnm{Chen}, \binits{C.-C.}},
\bauthor{\bsnm{Huang}, \binits{H.-H.}},
\bauthor{\bsnm{Chen}, \binits{H.-H.}}:
\bctitle{Ntusd-fin: a market sentiment dictionary for financial social media
  data applications}.
In: \bbtitle{Proceedings of the 1st Financial Narrative Processing Workshop
  (FNP 2018)},
pp. \bfpage{37}--\blpage{43}
(\byear{2018})
\end{bchapter}
\endbibitem

\bibitem{miller1998wordnet}
\begin{bbook}
\bauthor{\bsnm{Miller}, \binits{G.A.}}:
\bbtitle{WordNet: An Electronic Lexical Database},
(\byear{1998})
\end{bbook}
\endbibitem

\bibitem{loria2018textblob}
\begin{botherref}
\oauthor{\bsnm{Loria}, \binits{S.}}, et al.:
textblob documentation.
Release 0.15
\textbf{2}(8)
(2018)
\end{botherref}
\endbibitem

\bibitem{strapparava2007semeval}
\begin{bchapter}
\bauthor{\bsnm{Strapparava}, \binits{C.}},
\bauthor{\bsnm{Mihalcea}, \binits{R.}}:
\bctitle{Semeval-2007 task 14: Affective text}.
In: \bbtitle{Proceedings of the Fourth International Workshop on Semantic
  Evaluations (SemEval-2007)},
pp. \bfpage{70}--\blpage{74}
(\byear{2007})
\end{bchapter}
\endbibitem

\bibitem{goodfellow2014generative}
\begin{botherref}
\oauthor{\bsnm{Goodfellow}, \binits{I.}},
\oauthor{\bsnm{Pouget-Abadie}, \binits{J.}},
\oauthor{\bsnm{Mirza}, \binits{M.}},
\oauthor{\bsnm{Xu}, \binits{B.}},
\oauthor{\bsnm{Warde-Farley}, \binits{D.}},
\oauthor{\bsnm{Ozair}, \binits{S.}},
\oauthor{\bsnm{Courville}, \binits{A.}},
\oauthor{\bsnm{Bengio}, \binits{Y.}}:
Generative adversarial nets.
Advances in Neural Information Processing Systems
\textbf{27}
(2014)
\end{botherref}
\endbibitem

\bibitem{kritzman2010skulls}
\begin{barticle}
\bauthor{\bsnm{Kritzman}, \binits{M.}},
\bauthor{\bsnm{Li}, \binits{Y.}}:
\batitle{Skulls, financial turbulence, and risk management}.
\bjtitle{Financial Analysts Journal}
\bvolume{66}(\bissue{5}),
\bfpage{30}--\blpage{41}
(\byear{2010})
\end{barticle}
\endbibitem

\bibitem{whaley2009understanding}
\begin{barticle}
\bauthor{\bsnm{Whaley}, \binits{R.E.}}:
\batitle{Understanding the {VIX}}.
\bjtitle{The Journal of Portfolio Management}
\bvolume{35}(\bissue{3}),
\bfpage{98}--\blpage{105}
(\byear{2009})
\end{barticle}
\endbibitem

\bibitem{makoviychuk2021isaac}
\begin{botherref}
\oauthor{\bsnm{Makoviychuk}, \binits{V.}},
\oauthor{\bsnm{Wawrzyniak}, \binits{L.}},
\oauthor{\bsnm{Guo}, \binits{Y.}},
\oauthor{\bsnm{Lu}, \binits{M.}},
\oauthor{\bsnm{Storey}, \binits{K.}},
\oauthor{\bsnm{Macklin}, \binits{M.}},
\oauthor{\bsnm{Hoeller}, \binits{D.}},
\oauthor{\bsnm{Rudin}, \binits{N.}},
\oauthor{\bsnm{Allshire}, \binits{A.}},
\oauthor{\bsnm{Handa}, \binits{A.}},
\oauthor{\bsnm{State}, \binits{G.}}:
{Isaac Gym}: High performance {GPU}-based physics simulation for robot
  learning.
Datasets and Benchmarks Track, NeurIPS
(2021)
\end{botherref}
\endbibitem

\bibitem{Sharpe}
\begin{botherref}
\oauthor{\bsnm{Sharpe}, \binits{W.F.}}:
The {Sharpe} {Ratio}.
Journal of Portfolio Management
(1994)
\end{botherref}
\endbibitem

\bibitem{malkiel2003passive}
\begin{barticle}
\bauthor{\bsnm{Malkiel}, \binits{B.G.}}:
\batitle{Passive investment strategies and efficient markets}.
\bjtitle{European Financial Management}
\bvolume{9}(\bissue{1}),
\bfpage{1}--\blpage{10}
(\byear{2003})
\end{barticle}
\endbibitem

\bibitem{ang2012mean}
\begin{botherref}
\oauthor{\bsnm{Ang}, \binits{A.}}:
Mean-variance investing.
Columbia Business School Research Paper No. 12/49
(2012)
\end{botherref}
\endbibitem

\bibitem{finrl_podracer_2021}
\begin{botherref}
\oauthor{\bsnm{Li}, \binits{Z.}},
\oauthor{\bsnm{Liu}, \binits{X.-Y.}},
\oauthor{\bsnm{Zheng}, \binits{J.}},
\oauthor{\bsnm{Wang}, \binits{Z.}},
\oauthor{\bsnm{Walid}, \binits{A.}},
\oauthor{\bsnm{Guo}, \binits{J.}}:
{FinRL-Podracer}: High-performance and scalable deep reinforcement learning for
  quantitative finance.
ACM International Conference on AI in Finance (ICAIF)
(2021)
\end{botherref}
\endbibitem

\bibitem{liu2021podracer}
\begin{botherref}
\oauthor{\bsnm{Liu}, \binits{X.-Y.}},
\oauthor{\bsnm{Li}, \binits{Z.}},
\oauthor{\bsnm{Yang}, \binits{Z.}},
\oauthor{\bsnm{Zheng}, \binits{J.}},
\oauthor{\bsnm{Wang}, \binits{Z.}},
\oauthor{\bsnm{Walid}, \binits{A.}},
\oauthor{\bsnm{Guo}, \binits{J.}},
\oauthor{\bsnm{Jordan}, \binits{M.}}:
{ElegantRL-Podracer}: Scalable and elastic library for cloud-native deep
  reinforcement learning.
Deep Reinforcement Learning Workshop at NeurIPS
(2021)
\end{botherref}
\endbibitem

\bibitem{team2021open}
\begin{botherref}
\oauthor{\bsnm{Team}, \binits{O.E.L.}},
\oauthor{\bsnm{Stooke}, \binits{A.}},
\oauthor{\bsnm{Mahajan}, \binits{A.}},
\oauthor{\bsnm{Barros}, \binits{C.}},
\oauthor{\bsnm{Deck}, \binits{C.}},
\oauthor{\bsnm{Bauer}, \binits{J.}},
\oauthor{\bsnm{Sygnowski}, \binits{J.}},
\oauthor{\bsnm{Trebacz}, \binits{M.}},
\oauthor{\bsnm{Jaderberg}, \binits{M.}},
\oauthor{\bsnm{Mathieu}, \binits{M.}}, et al.:
Open-ended learning leads to generally capable agents.
arXiv preprint arXiv:2107.12808
(2021)
\end{botherref}
\endbibitem

\bibitem{akiba2019optuna}
\begin{botherref}
\oauthor{\bsnm{Akiba}, \binits{T.}},
\oauthor{\bsnm{Sano}, \binits{S.}},
\oauthor{\bsnm{Yanase}, \binits{T.}},
\oauthor{\bsnm{Ohta}, \binits{T.}},
\oauthor{\bsnm{Koyama}, \binits{M.}}:
Optuna: A next-generation hyperparameter optimization framework.
ACM SIGKDD International Conference on Knowledge Discovery \& Data Mining
(2019)
\end{botherref}
\endbibitem

\bibitem{liaw2018tune}
\begin{botherref}
\oauthor{\bsnm{Liaw}, \binits{R.}},
\oauthor{\bsnm{Liang}, \binits{E.}},
\oauthor{\bsnm{Nishihara}, \binits{R.}},
\oauthor{\bsnm{Moritz}, \binits{P.}},
\oauthor{\bsnm{Gonzalez}, \binits{J.E.}},
\oauthor{\bsnm{Stoica}, \binits{I.}}:
Tune: A research platform for distributed model selection and training.
ICML AutoML Workshop
(2018)
\end{botherref}
\endbibitem

\bibitem{gupta2021embodied}
\begin{botherref}
\oauthor{\bsnm{Gupta}, \binits{A.}},
\oauthor{\bsnm{Savarese}, \binits{S.}},
\oauthor{\bsnm{Ganguli}, \binits{S.}},
\oauthor{\bsnm{Fei-Fei}, \binits{L.}}:
Embodied intelligence via learning and evolution.
Nature Communications
(2021)
\end{botherref}
\endbibitem

\bibitem{scholl2021market}
\begin{botherref}
\oauthor{\bsnm{Scholl}, \binits{M.P.}},
\oauthor{\bsnm{Calinescu}, \binits{A.}},
\oauthor{\bsnm{Farmer}, \binits{J.D.}}:
How market ecology explains market malfunction.
Proceedings of the National Academy of Sciences
\textbf{118}(26)
(2021)
\end{botherref}
\endbibitem

\bibitem{guan2021explainable}
\begin{botherref}
\oauthor{\bsnm{Guan}, \binits{M.}},
\oauthor{\bsnm{Liu}, \binits{X.-Y.}}:
Explainable deep reinforcement learning for portfolio management: An empirical
  approach.
ACM International Conference on AI in Finance (ICAIF)
(2021)
\end{botherref}
\endbibitem

\bibitem{byrd2020differentially}
\begin{bchapter}
\bauthor{\bsnm{Byrd}, \binits{D.}},
\bauthor{\bsnm{Polychroniadou}, \binits{A.}}:
\bctitle{Differentially private secure multi-party computation for federated
  learning in financial applications}.
In: \bbtitle{Proceedings of the First ACM International Conference on AI in
  Finance},
pp. \bfpage{1}--\blpage{9}
(\byear{2020})
\end{bchapter}
\endbibitem

\bibitem{liu2021fate}
\begin{barticle}
\bauthor{\bsnm{Liu}, \binits{Y.}},
\bauthor{\bsnm{Fan}, \binits{T.}},
\bauthor{\bsnm{Chen}, \binits{T.}},
\bauthor{\bsnm{Xu}, \binits{Q.}},
\bauthor{\bsnm{Yang}, \binits{Q.}}:
\batitle{Fate: An industrial grade platform for collaborative learning with
  data protection.}
\bjtitle{Journal of Machine Learning Research}
\bvolume{22}(\bissue{226}),
\bfpage{1}--\blpage{6}
(\byear{2021})
\end{barticle}
\endbibitem

\bibitem{kairouz2021advances}
\begin{barticle}
\bauthor{\bsnm{Kairouz}, \binits{P.}},
\bauthor{\bsnm{McMahan}, \binits{H.B.}},
\bauthor{\bsnm{Avent}, \binits{B.}},
\bauthor{\bsnm{Bellet}, \binits{A.}},
\bauthor{\bsnm{Bennis}, \binits{M.}},
\bauthor{\bsnm{Bhagoji}, \binits{A.N.}},
\bauthor{\bsnm{Bonawitz}, \binits{K.}},
\bauthor{\bsnm{Charles}, \binits{Z.}},
\bauthor{\bsnm{Cormode}, \binits{G.}},
\bauthor{\bsnm{Cummings}, \binits{R.}}, \betal:
\batitle{Advances and open problems in federated learning}.
\bjtitle{Foundations and Trends{\textregistered} in Machine Learning}
\bvolume{14}(\bissue{1--2}),
\bfpage{1}--\blpage{210}
(\byear{2021})
\end{barticle}
\endbibitem

\bibitem{sutton2022quest}
\begin{botherref}
\oauthor{\bsnm{Sutton}, \binits{R.S.}}:
The quest for a common model of the intelligent decision maker.
arXiv preprint arXiv:2202.13252
(2022)
\end{botherref}
\endbibitem

\bibitem{DDPG}
\begin{botherref}
\oauthor{\bsnm{Lillicrap}, \binits{T.}},
\oauthor{\bsnm{Hunt}, \binits{J.}},
\oauthor{\bsnm{Pritzel}, \binits{A.}},
\oauthor{\bsnm{Heess}, \binits{N.}},
\oauthor{\bsnm{Erez}, \binits{T.}},
\oauthor{\bsnm{Tassa}, \binits{Y.}},
\oauthor{\bsnm{Silver}, \binits{D.}},
\oauthor{\bsnm{Wierstra}, \binits{D.}}:
Continuous control with deep reinforcement learning.
International Conference on Learning Representations (ICLR)
(2016)
\end{botherref}
\endbibitem

\bibitem{PPO_2017}
\begin{botherref}
\oauthor{\bsnm{Schulman}, \binits{J.}},
\oauthor{\bsnm{Wolski}, \binits{F.}},
\oauthor{\bsnm{Dhariwal}, \binits{P.}},
\oauthor{\bsnm{Radford}, \binits{A.}},
\oauthor{\bsnm{Klimov}, \binits{O.}}:
Proximal policy optimization algorithms.
arXiv:1707.06347
(2017)
\end{botherref}
\endbibitem

\bibitem{finrl_meta_2021}
\begin{botherref}
\oauthor{\bsnm{Liu}, \binits{X.-Y.}},
\oauthor{\bsnm{Rui}, \binits{J.}},
\oauthor{\bsnm{Gao}, \binits{J.}},
\oauthor{\bsnm{Yang}, \binits{L.}},
\oauthor{\bsnm{Yang}, \binits{H.}},
\oauthor{\bsnm{Wang}, \binits{Z.}},
\oauthor{\bsnm{Wang}, \binits{C.D.}},
\oauthor{\bsnm{Jian}, \binits{G.}}:
{FinRL-Meta}: Data-driven deep reinforcementlearning in quantitative finance.
Data-Centric AI Workshop, NeurIPS
(2021)
\end{botherref}
\endbibitem

\end{thebibliography}

\newpage

\newpage
\appendix

\definecolor{Gray}{RGB}{217,234,211}
\begin{table}
\caption{List of key terms for reinforcement learning.}
\small
\renewcommand{\arraystretch}{1.2}
\centering
\scalebox{0.75}{
\begin{tabular}{|l|l|m{200pt}<{\centering}|}
   \hline 
    \textbf{Key Terms} & \textbf{Description} \\
    \hline
    Agent \cite{sutton2022quest} 
    & A decision maker\\
    \hline
    Environment \cite{sutton2022quest}
    & A world with which an agent interacts with\\
    \hline
    Gym-style environment \cite{brockman2016openai}
    & A standard form of DRL environment by OpenAI \\
    \hline
    Markov Decision Process (MDP)
    & A mathematical framework to model decision-making problems\\
    \hline
    State, Action, Reward 
    & Three main factors in an agent-environment interaction \\
    \hline
    Policy
    & A rule that agent follow to make decision\\
    \hline
    Policy gradient & An approach to solve RL problems by optimizing the policy directly\\
    \hline
    Deep Q-Learning (DQN) \cite{DQN} &  The first DRL algorithm that uses a neural network to approximate the Q-function \\
    \hline
    DDPG \cite{DDPG} & Deep Deterministic Policy Gradient algorithm\\
    \hline
    PPO \cite{PPO_2017} & Proximal Policy Optimization algorithm\\
    \hline
    Hyperparameter tuning
    & Change hyperparameter during training to get a converged result faster \\
    \hline
    Ensemble strategy
    & An ML technique. Here we combine several DRL agents to a better model\\
    \hline
    Population-based training (PBT) & Optimise a population of models and hyperparameters, and select the optimal set\\
    \hline
    Generational evolution \cite{finrl_podracer_2021} & Employing an evolution strategy over generations \\
    \hline 
    Tournament-based evolution \cite{liu2021podracer} & An evolution by asynchronously updating a tournament board of models \\
    \hline
    Curriculum learning &  Training an ML model from easier to harder data, imitating the human curriculum\\
    \hline
    Simulation-to-reality gap 
    &  The difference between simulation environment and real-world task\\
    \hline
\end{tabular}}

\label{table:RL_terms}
\end{table}

\definecolor{Gray}{RGB}{217,234,211}
\begin{table}
\caption{List of key terms for finance.}
\small
\renewcommand{\arraystretch}{1.2} 
\centering
\scalebox{0.8}{
\begin{tabular}{|l|l|m{200pt}<{\centering}|}
   \hline 
    \textbf{Key Terms} & \textbf{Description} \\
    \hline
    Algorithmic trading & A method of trading using designed algorithm instead of human traders\\
    \hline
    Backtesting
    & A method to see how a strategy performs on a certain period of historical data\\
    \hline
    Signal-to-noise ratio (SNR)
    & A ratio of desired signal (good data) to undesired signal (noise)\\
    \hline
    DataOps & A series of principles and practices to improve the quality of data science\\
    \hline
    Sentiment data 
    & A category in financial big data that contains subjective viewpoints\\
    \hline
    Historical data
    & All kinds of data that already existed in the past\\
    \hline
    Survivorship bias
    & A bias caused by only seeing existing examples, but not those already died out\\
    \hline
    Information leakage
    & When the data contains future information, causing model overfitting\\
    \hline
    Paper trading 
    & Simulation of buying and selling without using real money\\
    \hline
    OHLCV
    & A popular form of market data with: Open, High, Low, Close, Volume\\
    \hline
    Technical indicators & A statistical calculation based on OHLCV data to indicate future price trends \\
    \hline
    Market frictions & A financial market friction as anything that interferes with trade.\\
    \hline
    Market crash
    & A huge drop of market price within a very short time\\
    \hline
    Volatility index (VIX) \cite{whaley2009understanding}
    & A market index that shows the market's expectations for volatility\\
    \hline
    Limit Order Book (LOB) & A list to record the interest of buyers and sellers\\
    \hline
    Smart beta index & An enhanced indexing strategy to beat a benchmark index\\
    \hline
    Liquidation, trade execution & An investor closes their position in an asset\\
    \hline
\end{tabular}}
\label{table:Finance_terms}
\end{table}


\section{Terminology}

We provide a list of key terms  for reinforcement learning and finance in Table~\ref{table:RL_terms} and Table~\ref{table:Finance_terms}. For terminologies of reinforcement learning, interested users can refer to \cite{sutton2022quest} or the classic textbook \cite{sutton2018reinforcement}. Also, the webpage\footnote{OpenAI SpinningUp: \url{https://spinningup.openai.com/en/latest/spinningup/rl_intro.html}} explains key concepts of RL. For terminologies of finance, interested users can refer to \cite{de2018advances}.

\newpage

\section{Dataset Documentation and Usages}

We organize the dataset documentation according to the suggested template of \textit{datasheets for datasets} \footnote{Timnit Gebru, Jamie Morgenstern, Briana Vecchione, Jennifer Wortman Vaughan, Hanna Wal- lach, Hal Daumé Iii, and Kate Crawford. Datasheets for datasets. \textit{Communications of the ACM}, 64(12):86–92, 2021.}.

\subsection{Motivation}
\begin{itemize}
    \item \textbf{For what purpose was the dataset created?}
    
    As data is refreshing minute-to-millisecond, finance is a particularly difficult playground for deep reinforcement learning.
    
    In academia, scholars use financial big data to obtain more complex and precise understanding of markets and economics. While industries use financial big data to refine their analytical strategies and strengthen their prediction models. To serve the rapidly growing AI4Finance community, we create FinRL-Meta that provides data access from different sources, pre-processes the raw data with different features, and builds the data to RL environments. We aim to provide dynamic RL environments that are manageable by users.
    
    We aim to build a financial metaverse, a universe of near real-market environments, as a playground for data-driven financial machine learning.
    
    \item \textbf{Who created the dataset?}
    
    FinRL-Meta is an open-source project created by the AI4Finance community. Contents of FinRL-Meta are contributed by the authors of this paper and will be maintained by members of the AI4Finance community.
    
    \item \textbf{Who funded the creation of the dataset?}
  
    AI4Finance Foundation, a non-profit open-source community that shares AI tools for finance, funded our project. 
\end{itemize}

\subsection{Composition}
\begin{itemize}
    \item \textbf{What do the instances that comprise the dataset represent?}
    
    Instances of FinRL-Meta are volume-price data includes: stocks, securities, cryptocurrencies, etc; and sentiment data from social media, ESG, Google Trends, etc. FinRL-Meta provides hundreds of market environments through an automatic pipeline that collects dynamic datasets from real-world markets and processes them into standard gym-style market environments. FinRL-Meta also benchmarks popular papers as stepping stones for users to design new trading strategies.
    
    \item \textbf{How many instances are there in total?}
    
    FinRL-Meta does not store data directly. Instead, we provide codes for a pipeline of data accessing, data cleaning, feature engineering, and building into RL environments. Table~\ref{tab:data_sources} provides the supported data sources of FinRL-Meta. 
    
    At the moment, there are hundreds of market environments, dozens of tutorials and demos, and several benchmarks provided.
    
    \item \textbf{Does the dataset contain all possible instances or is it a sample of instances from a larger set?}
    
    With our provided codes, users could fetch data from the data source by properly specifying the starting date, ending date, time granularity, asset set, attributes, etc. 
    
    \item \textbf{What data does each instance consist of?}
    
    Now there are several types of financial data, as shown in Table \ref{tab:data_sources}:
    
    \item \textbf{Is there a label or target associated with each instance?}
    
    No. There is not label or preset target for each instance. But users can use our benchmarks are baselines. 
    
    \item \textbf{Is any information missing from individual instances?}
    
    Yes. In several data sources, there are missing values and we provided standard preprocessing methods.
    
    \item \textbf{Are relationships between individual instances made explicit?}
    
    Yes. An instance is a sample set of the market of interest.
    
    \item \textbf{Are there recommended data splits?}
    
    We recommend users to follow our training-testing-training pipeline, as shown in Fig.~\ref{fig:finrl-meta timeline}. Users can flexibly choose their preferred settings, e.g., in stock trading task, our demo access Yahoo! Finance database and use data from 01/01/2009 to 06/30/2020 for training and data from 07/01/2020 to 05/31/2022 for backtesting.
    
    \item \textbf{Are there any errors, sources of noise, or redundancies in the dataset?}
    
    For the raw data fetched from different sources, there are noise and outliers. We provide codes to process the data and built them into standard RL gym environment.
    
    \item \textbf{Is the dataset self-contained, or does it link to or otherwise rely on external resources?}
    
    It is linked to external resources. As shown in Table~\ref{tab:data_sources}, FinRL-Meta fetch data from data sources to build gym environments.
    
    \item \textbf{Does the dataset contain data that might be considered confidential?}
    
    No. All our data are from publicly available data sources.
    
    \item \textbf{Does the dataset contain data that, if viewed directly, might be offensive, insulting, threatening, or might otherwise cause anxiety?}
    
    No. All our data are numerical.
    
\end{itemize}

\subsection{Collection Process}

\begin{itemize}
    \item \textbf{How was the data associated with each instance acquired?}
    
    FinRL-Meta fetches data from data sources. as shown in Table~\ref{tab:data_sources}.
    
    \item \textbf{What mechanisms or procedures were used to collect the data?}
    
    FinRL-Meta provides dynamic market environments that are built according to users' settings. To achieve this, we provide software APIs to fetch data from different data sources. Note that some data sources require accounts and passwords or have limitations on the number or frequency of requests.
    
    \item \textbf{If the dataset is a sample from a larger set, what was the sampling strategy?}
    
    It is dynamic, depending on users' settings, such as the starting date, ending date, time granularity, asset set, attributes, etc.
    
    \item \textbf{Who was involved in the data collection process and how were they compensated?}
    
    Our codes collect publicly available market data, which is free.
    
    
    \item \textbf{Over what timeframe was the data collected?}
    
    It is not applicable because the environments are created dynamically by running the codes to fetch data in real-time.
    
    \item \textbf{Were any ethical review processes conducted?}
    
    No?
    
\end{itemize}

\subsection{Preprocessing/cleaning/labeling}

\begin{itemize}
    \item \textbf{Was any preprocessing/cleaning/labeling of the data done?}
    
    Yes. For the raw data fetched from different sources, there are noise and outliers. We provide codes to process the data and built them into standard RL gym environment.
    
    \item \textbf{Was the “raw” data saved in addition to the preprocessed/cleaned/labeled data}
    
    The raw data are hold by different data sources (data providers).
    
    \item \textbf{Is the software that was used to preprocess/clean/label the data available?}
    
    Yes. We use our own codes to do cleaning and preprocessing.
    
\end{itemize}

\subsection{Uses}

\begin{itemize}
    \item \textbf{Has the dataset been used for any tasks already?}
    
    Yes. Thousands of AI4Finance community members use FinRL-Meta for learning and research purpose. There are also courses in colleges using FinRL-Meta as material for teaching financial reinforcement learning. Demos and tutorials are mentioned in Section \ref{benchmarks}.
    
    \item \textbf{Is there a repository that links to any or all papers or systems that use the dataset?}
    
    1. Research papers that used FinRL-Meta are listed here:
    
    \url{https://github.com/AI4Finance-Foundation/FinRL-Tutorials/blob/master/FinRL_papers.md}
    
    Our conference version of FinRL-Meta \cite{liu2022finrlmeta} appeared in NeurIPS 2022 Datasets and Benchmarks Track.
    
    Our workshop version of FinRL-Meta \cite{finrl_meta_2021} appeared in NeurIPS 2021 Workshop on Data-Centric AI.
    
    2. The following three repositories have incorporated FinRL-Meta:
    \begin{itemize}
        \item FinRL-Meta corresponding to the market layer of FinRL (7K stars):\\ \url{https://github.com/AI4Finance-Foundation/FinRL}
        \item ElegantRL (2.7K stars) supports FinRL-Meta:\\ \url{https://github.com/AI4Finance-Foundation/ElegantRL} 
        \item FinRL-Podracer: \\ \url{https://github.com/AI4Finance-Foundation/FinRL_Podracer}
    \end{itemize}
    
    \item \textbf{What (other) tasks could the dataset be used for?}
     
     Besides the current tasks (tutorial, demo and benchmarks), FinRL-Meta will be useful for the following tasks:
     \begin{itemize}
         \item \textbf{Curriculum learning for agents}: Based on FinRL-Meta (a universe of market environments, say $\geq 100$), one is able to construct an environment by sampling data samples from multiple market datasets, similar to XLand \cite{team2021open}. In this way, one can apply the curriculum learning method \cite{team2021open} to train a generally capable agent for several financial tasks.
         \item  To improve the performance for the large-scale markets, we are exploiting GPU-based massive parallel simulation such as Isaac Gym \cite{makoviychuk2021isaac}.
         \item It will be interesting to explore the evolutionary perspectives \cite{gupta2021embodied,scholl2021market,finrl_podracer_2021, liu2021podracer} to simulate the markets. We believe that FinRL-Meta will provide insights into complex market phenomena and offer guidance for financial regulations. 
     \end{itemize}
  
    \item \textbf{Is there anything about the composition of the dataset or the way it was collected and preprocessed/cleaned/labeled that might impact future uses?}
    
    We believe that FinRL-Meta will not encounter usage limits. Our data are fetched from different sources in real-time when running the codes. However, there may be one or two out of $\geq 30$ data sources (in Table \ref{tab:data_sources}) change data access rules that may impact future use. So please refer to the rules and accessibility of certain data sources when using.
     
    \item \textbf{Are there tasks for which the dataset should not be used?}
    
    No. Since there are no ethical problems for FinRL-Meta, users could use FinRL-Meta in any task as long as it does not violate laws.
    
    \textbf{Disclaimer: Nothing herein is financial advice, and NOT a recommendation to trade real money. Please use common sense and always first consult a professional before trading or investing.}
\end{itemize}

\subsection{Distribution}

\begin{itemize}
    \item \textbf{Will the dataset be distributed to third parties outside of the entity (e.g., company, institution, organization) on behalf of which the dataset was created?}
    
    No. It will always be held on GitHub under MIT license, for educational and research purposes.
    
    \item \textbf{How will the dataset be distributed?}
    
    Our codes and existing environments are available on GitHub FinRL-Meta repository \url{https://github.com/AI4Finance-Foundation/FinRL-Meta}.
    
    \item \textbf{When will the dataset be distributed?}
    
    FinRL-Meta is publicly available since February 14th, 2021.
    
    \item \textbf{Will the dataset be distributed under a copyright or other intellectual property (IP) license, and/or under applicable terms of use (ToU)?}
    
    FinRL-Meta is distributed under MIT License, for educational and research purposes.
    
    \item \textbf{Have any third parties imposed IP-based or other restrictions on the data associated with the instances?}
    
    No.
    
    \item \textbf{Do any export controls or other regulatory restrictions apply to the dataset or to individual instances?}
    
    No. Our data are fetched from different sources in real time. However, there may be one or two out of $\geq 20$ data sources (in Table \ref{tab:data_sources}) change data access rules that may impact future use. So please refer to the rules and accessibility of certain data sources when using.
\end{itemize}

\subsection{Maintenance}

\begin{itemize}
    \item \textbf{Who will be supporting/hosting/maintaining the dataset?}
    
    FinRL-Meta has been actively maintained by AI4Finance Foundation (including the authors of this paper) which has over 12K members at the moment (Mar. 2023). We are actively updating market environments, to serve the rapidly growing open-source AI4Finance community.
    
    \item \textbf{How can the owner/curator/manager of the dataset be contacted?}
    
    We encourage users to join 
    our Slack channel: \\
    \url{https://join.slack.com/t/ai4financeworkspace/shared_invite/zt-v670l1jm-dzTgIT9fHZIjjrqprrY0kg} \\
    or our mailing list:\\ \url{https://groups.google.com/u/1/g/ai4finance}, \\
    
    \item \textbf{Is there an erratum?}
    
    Users can use GitHub to report issues/bugs and use Slack channel, Discord channel, or mailing list (AI4Finance\_FinRL at \url{https://groups.google.com/u/2/g/ai4finance}) to discuss solutions. AI4Finance community is actively improving the codes, say extracting technical indicators, evaluating feature importance, quantifying the probability of model overfitting, etc.
    
    \item \textbf{Will the dataset be updated?}
    
    Yes, we are actively updating codes and adding more data sources. Users could get information and the newly updated version through our GitHub repository, or join the mailing list: \url{https://groups.google.com/u/1/g/ai4finance}.
    
    \item \textbf{If the dataset relates to people, are there applicable limits on the retention of the data associated with the instances}
    
    The data of FinRL-Meta do not relate to people.
    
    \item \textbf{Will older versions of the dataset continue to be supported/hosted/maintained?}
    
    Yes. All versions can be found on our GitHub repository.
    
    \item \textbf{If others want to extend/augment/build on/contribute to the dataset, is there a mechanism for them to do so?}
    
    We maintain FinRL-Meta on GitHub. Users can use GitHub to report issues/bugs and use Slack channel or mailing list to discuss solutions. We welcome community members to submit pull requests through GitHub.
    
\end{itemize}

\end{document}